\documentclass[journal,twocolumn]{IEEEtran}
\ifCLASSINFOpdf
\else
\fi

\usepackage[ruled,vlined]{algorithm2e}
\usepackage{threeparttable}
\usepackage{graphicx}
\usepackage{multirow}
\usepackage{hhline}
\usepackage{threeparttable}
\usepackage{multirow}
\usepackage{xcolor}
\usepackage{booktabs}
\usepackage{multicol}
\usepackage{graphicx}
\usepackage{float}
\usepackage{subfig}
\usepackage{makecell}
\usepackage{array}
\usepackage{multirow}
\usepackage{longtable}
\usepackage{rotating}
\usepackage{amsmath}
\usepackage{colortbl}
\usepackage{amssymb}
\usepackage{setspace}
\usepackage{hyperref}

\begin{document}

\title{A Reinforcement Learning-assisted Genetic Programming Algorithm for Team Formation Problem Considering Person-Job Matching}
%
%
%
\author{Yangyang Guo, Hao Wang*, Lei He, Witold Pedrycz,~\IEEEmembership{Life~Fellow,~IEEE},\\
	P.  N. Suganthan,~\IEEEmembership{Fellow,~IEEE}, Yanjie Song*
	\thanks{* Corresponding author: Hao Wang, Yanjie Song}
	\thanks{Yangyang Guo is with the College of Systems Engineering, National University of Defense Technology, China, and the School of Transportation Engineering, Dalian Maritime University, China.
	Hao Wang, Lei He, and Yanjie Song are with the College of Systems Engineering, National University of Defense Technology, China.
	Witold Pedrycz is with the Department of Electrical and Computer Engineering, University of Alberta, Canada. 
	P. N. Suganthan is with the KINDI Center for Computing Research, College of Engineering, Qatar University, Qatar, and the School of Electrical and Electronic Engineering, Nanyang Technological University, Singapore.}
	\thanks{$ ^1 $ Lei He and Yangyang Guo contributed equally to this article.}
	
}
	
	
	%
	%


\maketitle

\begin{abstract}
	An efficient team is essential for the company to successfully complete new projects. To solve the team formation problem considering person-job matching (TFP-PJM), a 0-1 integer programming model is constructed, which considers both person-job matching and team members' willingness to communicate on team efficiency, with the person-job matching score calculated using intuitionistic fuzzy numbers. Then,  a reinforcement learning-assisted genetic programming algorithm (RL-GP) is proposed to enhance the quality of solutions. The RL-GP adopts the ensemble population strategies. Before the population evolution at each generation, the agent selects one from four population search modes according to the information obtained, thus realizing a sound balance of exploration and exploitation. In addition, surrogate models are used in the algorithm to evaluate the formation plans generated by individuals, which speeds up the algorithm learning process. Afterward, a series of comparison experiments are conducted to verify the overall performance of RL-GP and the effectiveness of the improved strategies within the algorithm. The hyper-heuristic rules obtained through efficient learning can be utilized as decision-making aids when forming project teams. This study reveals the advantages of reinforcement learning methods, ensemble strategies, and the surrogate model applied to the GP framework. The diversity and intelligent selection of search patterns along with fast adaptation evaluation, are distinct features that enable RL-GP to be deployed in real-world enterprise environments.
\end{abstract}

\begin{IEEEkeywords}
	team formation, reinforcement learning, genetic programming, intuitionistic fuzzy numbers, ensemble population strategy, surrogate model
\end{IEEEkeywords}

%
\IEEEpeerreviewmaketitle

\section{Introduction}

\IEEEPARstart{I}{n} recent years, the development of numerous modern enterprises has been characterized by being technology-intensive, with talents playing a crucial role in facilitating their growth and success. However, many enterprises fail to optimize their talent resources when formulating development strategies, often neglecting the implementation of human resource mechanisms such as recruitment and training. This has resulted in a lack of responsiveness to market demand, hindering their progress. Therefore, it is crucial for enterprises to recognize the importance of human resources in the development of their organizations, as they are core resources that significantly impact competitiveness \cite{liu2021data}. A focus on talent management can assist enterprises in navigating complex social environments and fierce market competition, ultimately safeguarding the economic benefits of the enterprise.

The process of selecting and allocating candidates to various job positions within a team is a crucial aspect of human resource management. This problem can be called the team formation problem considering person-job matching (TFP-PJM). In the job matching problem, decision-makers need to scientifically allocate candidates to suitable positions according to the available information, leveraging the value of candidates to maximize the match between the two. Effective matching of candidates with suitable job positions can enhance job satisfaction and productivity, leading to higher organizational performance. Conversely, incorrect allocation of candidates can cause the loss of talent and significant damage to the enterprise \cite{he2021self}. In addition, many companies need to form a team to complete a specific project, and the candidates in the team must cooperate with each other. There may be differences in the communication willingness among different candidates, which can affect team efficiency. Therefore, to ensure the synergistic efficiency of the team, decision-makers need to consider how to select a certain number of candidates from a pool of candidates, which not only meets the skills required to build the team but also maximizes the degree of collaboration between candidates.

Team formation is an indispensable part of the modern organization, and many scholars have studied it from different aspects. The literature \cite{juarez2021comprehensive} provides a comprehensive overview of the studies conducted on team formation problems and groups them into two categories: Assignment-based team formation and community-based team formation. In these studies, objective functions such as maximizing a cricket team's scoring performance \cite{ahmed2013multi}, maximizing the team gains \cite{abdelsalam2009multi}, minimizing communication costs\cite{niveditha2017genetic}, etc., are considered. Many studies have different methods for quantifying skill scores, including precise number score \cite{perez2019players}, the intuitionistic fuzzy number \cite{okimoto2016mission}, probability variables \cite{liemhetcharat2012modeling}, etc. However, few studies have focused on the effects of both person-job matching and personnel relations on team efficiency.

The team formation problem is NP-hard \cite{juarez2021comprehensive}, and researchers have used multiple heuristics or meta-heuristics to solve this problem, such as genetic algorithm \cite{wi2009team}, particle swarm algorithm \cite{abdelsalam2009multi}, \cite{zhang2013multi}, and local search algorithm \cite{gutierrez2016multiple}. There are also hybrid algorithms that combine several search strategies or combine several algorithms with each other, focusing on global and local search performance, achieving the effect of constantly optimizing the formation scheme. However, these algorithms still have some problems such as low search efficiency and high dependence on algorithm parameters. Additionally, due to the multiple information about candidates and positions involved in the TFP-PJM problem, it is difficult to represent all properties of the problem using a simple rule. Artificially set heuristic rules may not be readily adjusted, which may ignore important properties that affect the performance of the algorithm and the quality of the search solution.

In contrast, genetic programming (GP), as a hyper-heuristic, can make full use of the characteristics of candidates and positions in the TFP-PJM problem, and combine them within specific expressions. During the population evolution, each individual will combine different low-level rules to form a hyper-heuristic rule to obtain the full ranking of candidates and generate a team formation scheme. In addition, many studies have also explored the integration of genetic programming with multiple strategies to solve various combinatorial optimization problems, such as the job shop scheduling problem \cite{zhang2021multitask}, the uncertain capacitated arc routing problem \cite{ardeh2021genetic}, and the online resource allocation problem \cite{tan2020cooperative}. These successful experiences can guide future algorithm design. However, traditional genetic programming is prone to the defect of a single search mode, which can overemphasize the global search and cannot effectively mine the local solution space. Moreover, when applying GP to solve combinatorial optimization problems like TFP-PJM, it is necessary to evaluate the generated formation schemes of each individual and check various types of constraints, which leads to the high computational cost of fitness evaluation. Therefore, to address the above problems, we improve the traditional GP and propose a reinforcement learning-assisted genetic programming algorithm (RL-GP) to solve the TFP-PJM problem. First, a multi-population ensemble strategy is introduced into the design of the algorithm, and different search modes are selected to assist population search by reinforcement learning, breaking the limitation of a single search mode of GP. A kind of $\varepsilon$-greedy strategy is used for agent action selection, whereby the action with the largest Q-value or one of the actions is randomly chosen based on the relationship between the random number and the threshold. Furthermore, we adopt agent modeling to quickly compute the fitness function values using a machine learning model, thereby reducing computational costs and enhancing algorithm learning. To improve the accuracy of the surrogate model, the real model is utilized to assess the individual fitness function values after a certain number of generations, and the surrogate model is updated online. The main contributions of this study are as follows:

1. A 0-1 integer programming model is constructed for the team formation problem considering person-job matching. Since the problem of whether a person is matched with a job is a fuzzy concept, it is suitable to use fuzzy set-oriented mathematical tools to deal with such a problem. In the mathematical model, intuitionistic fuzzy numbers are used to evaluate the personnel to obtain the person-job matching degree, and the communication matrix is used to calculate the cooperation degree among the selected personnel. On this basis, an optimization function aiming at maximizing team efficiency is proposed. Moreover, the mathematical model also accounts for constraints such as position restrictions and personnel ability constraints.

2. A reinforcement learning-assisted genetic programming algorithm (RL-GP) is proposed. The ensemble population strategies are used in the algorithm, and search modes are selected from the ensemble strategies for population search by reinforcement learning. In addition, the RL-GP algorithm introduces the surrogate model to expedite fitness computation and improve algorithmic learning efficiency.

3. Extensive experiments demonstrate the effectiveness of the improved strategies in the RL-GP algorithm to perform efficient learning and obtain hyper-heuristic rules superior to the construction heuristic algorithms. In addition, by comparing the quality of solutions generated by other search algorithms, it is proved that the RL-GP algorithm has advantages in solving the TFP-PJM problem. Experiments also indicate the benefit of the RL-assisted search mode selection strategy to take advantage of population search.

The paper is structured as follows: Section II presents related work on the team formation problem and genetic programming in solving the combinatorial optimization problem. Section III describes the TFP-PJM problem and outlines the model construction process. Section IV presents the details related to the reinforcement learning-assisted genetic programming algorithm to address the TFP-PJM problem. Section V verifies the effectiveness of the proposed algorithm through experiments. In the last section, we summarize the main work and propose future research directions.

\section{Related Work}
This section introduces studies related to the team formation problem and genetic programming in solving combinatorial optimization problems.
\subsection{Team Formation Problem}

Currently, the team formation problem has attracted a lot of attention and many scholars have studied the related problems from different angles, proposed various methods to construct models, and used different algorithms to solve the problem. Juárez, J et al.\cite{juarez2021comprehensive} proposed that model building can be divided into two types, namely assignment-based team formation and community-based team formation. The model proposed by Bart H. Boon et al.\cite{boon2003team} can be used for soccer teams to select candidates to form an optimal team and can calculate the value of the newly added players to the team. Fitzpatrick, E. L et al.\cite{fitzpatrick2005forming} modeled individual instinctive tendencies as performance indicators for team selection based on the assurance that team members' skills meet the requirements. Xu et al.\cite{xu2012modeling} formed a team based on three important characteristics, which are the professional knowledge level of candidates in related fields, the diversity of skills, and the willingness to cooperate among candidates. Baykasoglu, A et al.\cite{baykasoglu2007project} calculated the skilled fitness of candidates by fuzzy scoring method, and proposed a fuzzy multi-objective optimization model considering the time and budget constraints of projects and interpersonal relations among candidates. Liemhetcharat, S et al.\cite{liemhetcharat2012modeling} transformed team formation into agents with different skills collaborating with each other to complete a task, and proposed a learning algorithm that can learn good synergy graphs to form teams without prior knowledge.

Many scholars have studied the effect of interpersonal relationships on team efficiency and proposed different modeling approaches. Wang et.al.\cite{wang2015comparative} calculated the cost of communication between candidates through social network graphs and provided a public platform to implement several algorithms for team formation. Gutiérrez, J. H et al.\cite{gutierrez2016multiple} used a communication matrix to set up three possible scenarios of willingness to cooperate between candidates, and studied the efficiency formula for multi-project teams, which can provide help to solve many real-world cases.

Many algorithms have been used to solve the team formation problem. Zhang et al. \cite{zhang2013multi} conducted an evaluation of personnel cooperation using the MBTI measurement tool and employed a multi-objective particle swarm optimization algorithm to achieve a Pareto solution of team formation alternatives. Strnad, D et al.\cite{strnad2010fuzzy} developed a fuzzy genetic analysis model, which combined multiple criteria to define a single composite objective function, and proposed an island genetic algorithm with mixed crossover to optimize the team formation scheme. Bhowmik, A et al.\cite{bhowmik2014submodularity} modeled the team formation problem as an unconstrained submodular function maximization problem, and proposed a stochastic approximation scheme based on simulated annealing.

\subsection{Genetic Programming for Solving Combinatorial Optimization Problems}
In this paper, we use genetic programming to generate a team formation scheme. GP has the advantage of being a hyper-heuristic algorithm capable of accommodating a wide range of state characteristics associated with different combinatorial optimization problems. Xu et al.\cite{xu2021genetic} utilized the genetic programming algorithm to make routing and sequencing decisions that incorporated three delayed routing strategies to address the dynamic flexible job shop scheduling problem. Wang et al.\cite{wang2021genetic} addressed the uncertain capacitated arc routing problem by employing the genetic programming algorithm and introducing a niching technique to simplify routing policies, which improved algorithmic efficiency and performance.

In addition, GP has the ability to integrate multiple strategies to enhance algorithmic efficiency and quality, which has been applied in several fields. Zhang et al.\cite{zhang2021surrogate} combined genetic programming with a surrogate model to solve the dynamic flexible job shop scheduling problem, which improved the quality of the scheduling heuristic algorithm by making full use of the knowledge learned from different scheduling tasks. Ardeh et al.\cite{ardeh2021genetic} investigated the uncertain capacitated arc routing problem and introduced knowledge transfer through genetic programming, leading to an improvement in the algorithm search efficiency by limiting the exploration to unexplored domains.

To summarize, when studying the team formation problem, few studies have quantified both the two metrics of person-job matching and the level of cooperation between team members. This paper seeks to address this gap by simultaneously considering the impact of these two dimensions on team efficiency, in order to enhance the accuracy of the model and solution scheme. Many algorithms for solving team formation problems require the manual design of heuristic rules, which has certain limitations. While GP has the advantage of being able to automatically select low-level rules for combination, thereby reducing the need for specialized domain knowledge. Additionally, no research has been conducted on the combination of genetic programming and reinforcement learning to solve the problem.

\section{Model}

\label{Model}

This section describes the TFP-PJM problem in detail and introduces the variables and symbols, objective functions, and the setting of constraints involved in the model.

\subsection{Problem Description}
How to assemble a team is an important decision for any company when embarking on a new project. There are various factors that should be considered when making the decision. First, the skill match degree of the candidate is an important factor, because, in the event that an employee lacks the necessary skills, the project may be delayed or even rendered impossible to complete. There is a need to select those from the candidate pool whose skills match the requirements and who are capable of performing the job. Second, effective communication and collaboration are essential elements for achieving success in a team-based project. The efficacy of communication among team members has a direct impact on the project's timeline and quality. Consequently, careful consideration should be given to selecting candidates with strong communication skills to ensure smooth partnerships with other team members \cite{farasat2016social}. Overall, only with a combination of many considerations can a company assemble an efficient and collaborative team that provides strong support for the project's success \cite{fathian2017new}.

\begin{figure}[htp]
	\centering
	\includegraphics[width=0.45\textwidth]{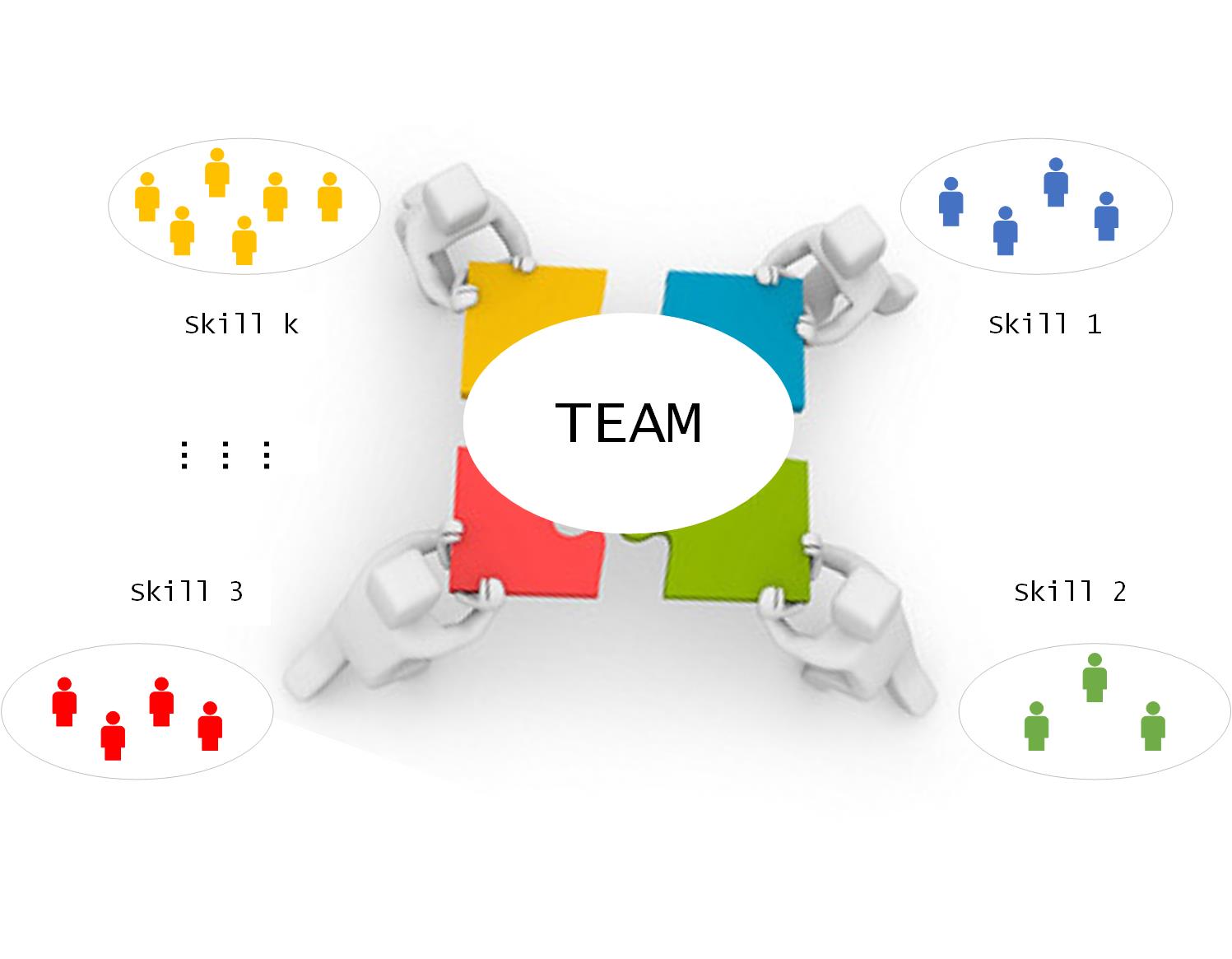}
	\caption{A team of members with different skills}
	\label{A team of members with different skills}
\end{figure}

When a company needs to undertake a new project, the TFP-PJM problem emphasizes the importance of selecting candidates with different skills and placing them in appropriate positions that meet the skill requirements of their respective positions. The assembled team must work together in a harmonious atmosphere to ensure optimal productivity. Skill assessment accuracy is also important in the TFP-PJM problem. Some skills may be overemphasized, while others may be overlooked. Therefore, companies need to clarify the weight of skills required for different positions based on project needs and job content, and then assess the candidates' skills on this basis. In addition, as shown in Figure \ref{A team of members with different skills}, certain tasks may necessitate the involvement of multiple members with similar skills, and the skills possessed by different candidates may overlap, just to be able to solve this problem. However, due to various factors such as personality and cultural differences, the willingness to communicate among candidates is different. ln general, there are candidates who possess a cooperative mindset, those who exhibit a neutral attitude, and those who are less inclined to cooperate \cite{gutierrez2016multiple}. When team members possess a strong willingness to communicate, the team's overall efficiency naturally increases.

What needs to be decided in the model is how to select candidates from the available pool and assign them to various team positions. This decision can not only meet the skill requirements of different positions but also meet the skill requirements set by the new project. At the same time, members can cooperate efficiently to maximize the team's efficiency.

\subsection{Variables and Symbols}
$C$: the set of candidates, $C=\{c_1,c_2,...,c_{|C|}\}$, $c_i$ stands for the candidate $i$;

$S$: the set of skills for all candidates, $S=\{s_1,s_2,...,s_{|S|}\}$, $s_k$ stands for the skill $k$;

$T$: the set of skills required for the task, $T=\{s_m,...,s_n\}\subseteq S$;

$S(c_i)$: the set of skills possessed by candidate $i$, $S(c_i)\subset S$;

$TS(c_{i})$: the set of skills required for task $T$, which is possessed by candidate $i$, $TS(c_{i})\subset S$;

$C(s_k)$: the set of candidates with skill $k$, $C(s_k)\subset V$;

$n(s_k)$: the required number of members with skill $k$ to complete task $T$;

$POS$: the set of positions in the team, $POS=\{pos_1,pos_2,...,pos_{|POS|}\}$, $pos_j$ stands for the position $j$;

$RS(pos_j)$: the set of skills required for position $j$, $RS(pos_j)=\{s_y,...,s_z\}$, $RS(pos_j)\subseteq S$;

$W$: the position assessment competency weight matrix, $w_{jl}$ indicates the assessment weight of position $j$ for the $l$th competence dimension, $\sum_{l=1}^{|L|}w_{jl}=1,\forall j=1,2,...,|POS|$;

$A$: the set of intuitionistic fuzzy numbers, $A=\{A_1,A_2\}$, $A_1\cup A_2=A$, $A_1\cap A_2=\emptyset$, $A=\{\left(\mu_{ijl},v_{ijl}\right)\Big|i=1,2,...,|C|;j=1,2,...,|POS|;l=1,2,...,|L|\}$, $A_1=\{\mu_{ijl}\big|i=1,2,...,|C|;j=1,2,...,|POS|;l=1,2,...,|L|\}$, $A_2=\{v_{ijl}\big|i=1,2,...,|C|;j=1,2,...,|POS|;l=1,2,...,|L|\}$, $\alpha_{ijl}=<\mu_{ijl},v_{ijl}>$, $\alpha_{ijl}$ denotes the intuitive fuzzy number score of candidate $i$ on the $l$th competence dimension of position $j$, where $\mu_{ijl},v_{ijl}$ stands for membership and non-membership respectively;

$ez_{ij}$: the match score of candidate $i$ with position $j$;

$R$: the personnel relationship matrix, the size of the matrix is $|C|\times|C|$, $r_{ii^{'}}$ indicates the willingness of candidates $i$ and $i^{'}$ to communicate. The value of $r_{ii^{'}}$ is shown as follows:

\begin{equation}
	r_{ii}^{'}=\left\{\begin{array}{cc}-1&\text{unwilling to cooperate}\\ 0&\text{neutral}\\ 1&\text{willing to cooperate}\end{array}\right.
\end{equation}

$\gamma$: the communication efficiency of the team.

\textbf {Decision Variables:}

$x_{ij}$: whether candidate $i$ is assigned to position $j$. The value of $x_{ij}$ is shown as follows:
\begin{equation}
	x_{ij}=\left\{\begin{matrix}1&\text{if the candidate is chosen}\\ 0&\text{else}\end{matrix}\right.
\end{equation}

\subsection{Model}
To better focus on the characteristics of the team formation problem itself, we make the following assumptions:

\textbf {Assumptions:}

1. Each candidate has at least one skill required by the team, and all candidates are likely to be selected.

2. There are more candidates than required team members.

3. Once the team is formed, members will not leave or be replaced during the project.

4. The skills required for the team and each position will not change.

5. The number of team positions is determined before selecting candidates, and no additional positions will be added or eliminated.

6. Each candidate can only use their specific skills to complete tasks.

7. The willingness of candidates to communicate with each other is predetermined and remains consistent throughout the project.

8. Some skills required by the team need multiple people with these skills to work together, while all skills required by each position need only one person to complete \cite{gutierrez2016multiple}.

The traditional person-job matching evaluation methods generally use precise numbers to express the indicators that affect the matching results, but due to the complexity and uncertainty of the actual matching problems, it is difficult to quantify the index values accurately. In order to arrange the selected candidates to the most suitable positions and give full play to their respective abilities, namely to achieve the maximum person-job matching, the intuitionistic fuzzy set is introduced into the model construction. First, we introduce the concept and definition of IFS:

Assuming that $X$ is a non-empty classical set, the IFS $A$ defined on $X$ can be expressed as:

\begin{equation}
	A=\{<x,\mu_{\mathrm{A}}(x),v_{\mathrm{A}}\left(x\right),{h_{\mathrm{A}}}\left(x\right)>|x\in X\}
\end{equation}
where $\mu_{A}(x)$, $v_{A}(x)$ and $h_{A}(x)$ denotes the membership, non-membership and hesitant functions respectively \cite{xu2007intuitionistic}. IFN is the essential element of the IFS theory, which can be easier expressed as $\alpha=\big\langle\mu,v\big\rangle$, and $\mu,v,\mu+v\in[0,1]$.

The model introduces the IFNs into the process of candidate evaluation. Under the evaluation dimension of each position, decision-makers make 0-1 decisions to judge whether a candidate can meet the requirements of a specific position or they can choose to abstain. After the evaluation, the intuitionistic fuzzy matrix of multiple decision-makers will be obtained. Since the membership function maybe 0 in the decision-making process, refer to \cite{he2016intuitionistic}, the intuitionistic fuzzy power interactive weighted average (IFPIWA) operator is used in the model to integrate the fuzzy information by setting different weights to each position's assessment dimensions to obtain the aggregated IFNs and the person-job matching score. The specific calculation formulas are shown as follows:

\begin{equation}
	d(\alpha_{ijl},\alpha_{ijl^{\prime}})=\frac{1}{2}(|\mu_{ijl}-\mu_{ijl'}|+|\text{v}_{ijl}-\text{v}_{ijl'}|)
\end{equation}
\begin{equation}
	\sup\left(\alpha_{ijl},\alpha_{ijl'}\right)=1-d(\alpha_{ijl},\alpha_{ijl'})
\end{equation}
\begin{equation}
	T(\alpha_{ijl})=\sum_{l^{'}=1,l^{'}\neq l}^{|L|}w_{jl^{'}}\sup\left(\alpha_{ijl^{'}},\alpha_{ijl}\right),l=1,2,...,|L|
\end{equation}

\begin{equation}
	\rho_{ijl}=\frac{w_{jl}(1+T(\alpha_{ijl}))}{\sum_{l=1}^{\lvert L\rvert}w_{jl}(\lvert1+T(\alpha_{ijl}))}
\end{equation}

\begin{figure*}[!t]
	\begin{equation}
		\begin{aligned}
			\text{IFPIWA}(\alpha_{ij1},\alpha_{ij2},\ldots,\alpha_{ij|L|})=\left(1-\prod_{l=1}^{|L|}\left(1-\mu_{ijl}\right)^{\rho_{ijl}},\prod_{l=1}^{|L|}(1-\mu_{ijl})^{\rho_{ijl}}-\prod_{l=1}^{|L|}\left(1-(\mu_{ijl}+v_{ijl})\right)^{\rho_{iji}}\right)
		\end{aligned}
	\end{equation}
\end{figure*}

\begin{figure}[htp]
	\centering
	\includegraphics[width=0.45\textwidth]{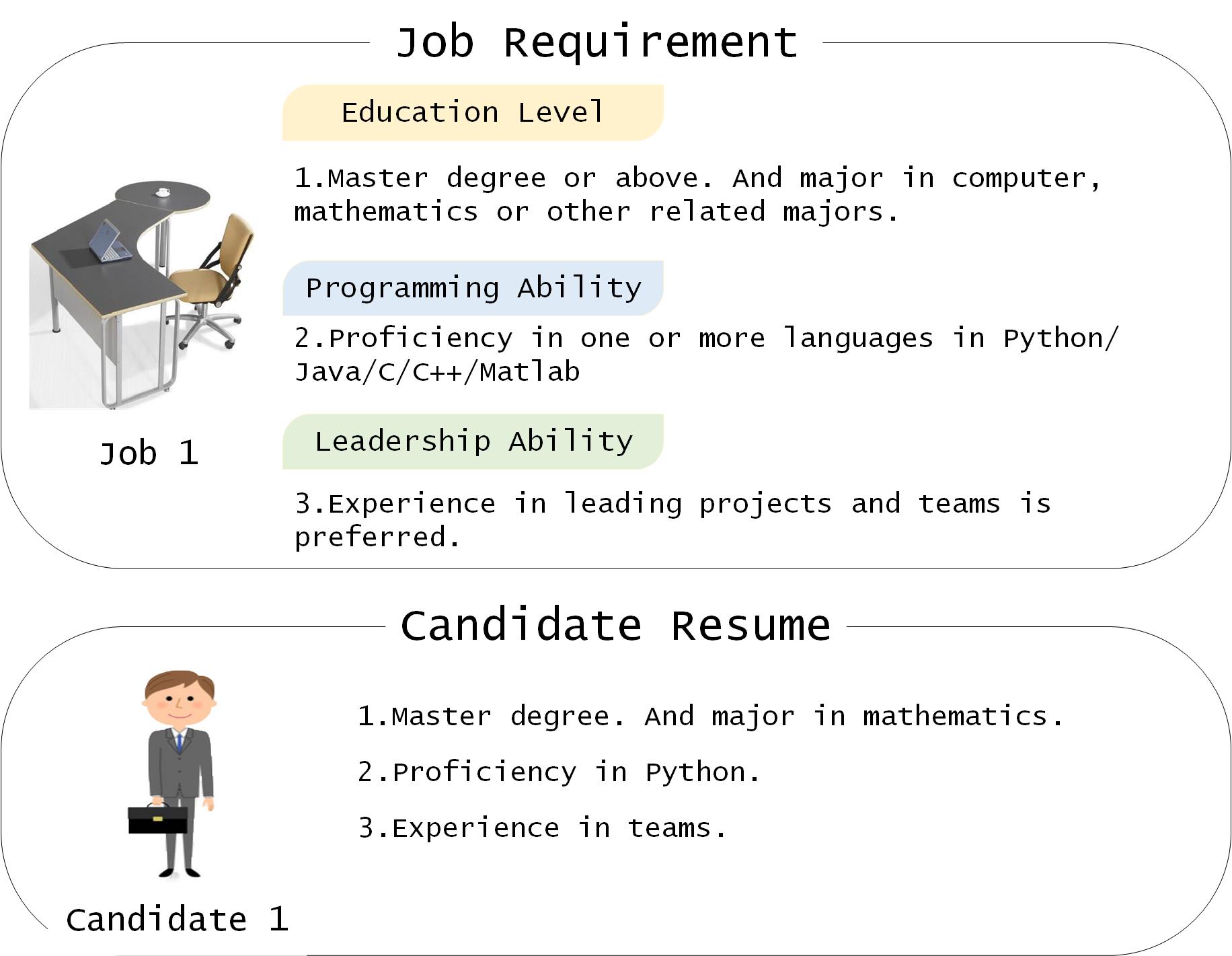}
	\caption{A person-job fit example}
	\label{A person-job fit example}
\end{figure}

Here, we give an example to facilitate the understanding of the definition. Figure \ref{A person-job fit example} shows the skill requirements of position $pos_1$ and the skill attributes possessed by candidate $c_1$. Suppose that three dimensions are set for the position, namely education level, programming ability, and leadership ability, with weights of 0.4,0.35 and 0.25 respectively, corresponding to the values of $w_{11}$, $w_{12}$, $w_{13}$. After scoring by experts, the candidate's intuitive fuzzy numbers in each dimension of the position will be obtained, which are $\alpha_{111}=<0.4,0.6>,\alpha_{112}=<0.6,0.4>,\alpha_{113}=<0.8,0.2>$. Through the information, the person-job matching score can be calculated. Firstly, the intuitionistic fuzzy Hamming distance of the three IFNs can be calculated.
\begin{equation}
	d(\alpha_{111},\alpha_{112})=\frac{1}{2}(|0.4-0.6|+|0.6-0.4|)=0.2
\end{equation}
\begin{equation}
	d(\alpha_{111},\alpha_{113})=\frac{1}{2}(|0.4-0.8|+|0.6-0.2|)=0.4
\end{equation}
\begin{equation}
	d(\alpha_{112},\alpha_{113})=\frac{1}{2}(|0.6-0.8|+|0.4-0.2|)=0.2
\end{equation}

Then, calculate the support between them.
\begin{equation}
	\sup{(\alpha_{111},\alpha_{112})}=1-0.2=0.8
\end{equation}
\begin{equation}
	\sup{(\alpha_{111}},\alpha_{113})=0.6,\sup{(\alpha_{112},\alpha_{113})}=0.8
\end{equation}
\begin{equation}
	T(\alpha_{111})=0.43,T(\alpha_{112})=T(\alpha_{113})=0.52
\end{equation}
\begin{equation}
	\rho_{111}=\frac{w_{11}(1+T(\alpha_{111}))}{\sum_{l=1}^{3}w_{1l}(1+T(\alpha_{11l}))}=0.39
\end{equation}
\begin{equation}
	\rho_{112}=\frac{w_{12}(1+T(\alpha_{112}))}{\sum_{l=1}^{3}w_{1l}(1+T(\alpha_{11l}))}=0.36
\end{equation}
\begin{equation}
	\rho_{113}=\frac{w_{13}(1+T(\alpha_{113}))}{\sum_{l=1}^{3}w_{1l}(1+T(\alpha_{11l}))}=0.26
\end{equation}
\begin{equation}
	\quad\text{IFIWA}(\alpha_{111},\alpha_{112},\alpha_{113})=<0.61,0.39>
\end{equation}

Finally, the matching score between the candidate $c_1$ and the position $pos_1$ is calculated to be $ez_{11}=0.22$.

In addition, the efficacy of communication within the team is influenced by the willingness to communicate among the members. Refer to \cite{gutierrez2016multiple}, the communication matrix is used to express the levels of cooperation among members. This matrix comprises values of 1, 0, or -1, indicating that two people are willing to collaborate, have a neutral attitude, or are not inclined to cooperate, respectively. A formulation for the overall communication efficiency of the team is designed based on the communication matrix, which is calculated as follows:
\begin{equation}
	\gamma=a\left(b+\frac{\sum_{i,i'\in C}r_{ii'}\cdot x_{ij}\cdot x_{i'j'}}{[\sum_{s_k\in T}n(s_k)]^2}\right)
\end{equation}

The communication efficiency of a team is obtained by summing up the communication willingness of all team members. If the candidate $i$ and $i'$ are willing to cooperate, and both are selected into the team and assigned to different positions, the numerator in the formula will increase by 1, thus the communication efficiency of the team will be improved. In this problem, we set $a=\frac{1}{2},b=3$, converting the interval of $\gamma$ from $ \left[-1,1\right] $ to $ \left[1,2\right] $, which is convenient for the subsequent setting of the objective function \cite{gutierrez2016multiple}.

The goal of the model is to maximize team productivity by carefully selecting suitable candidates and placing them in the right positions. This requires considering both the person-job matching score and the communication efficiency score, and the objective function is shown below:

\textbf {Objective Function:}
\begin{equation}
	\quad max\sum_{i\in C}\sum_{j\in POS}ez_{ij}\cdot x_{ij}\cdot\gamma
\end{equation}
where $ez_{ij}$ denotes the match score of the selected candidates and their respective positions, and $\gamma$ denotes the communication efficiency of the personnel in the team, which are two important factors affecting the efficiency of the whole team. The interval value of $\gamma$ makes it possible to multiply the team communication efficiency and the person-job matching score to express the team productivity. Specifically, the larger the calculated value of communication efficiency, the larger the value of team productivity obtained by multiplying it with the person-job matching score.

The model mainly considers the job constraints and the skill constraints of the personnel, which are expressed as follows:

\textbf {Constraints:}

1. Each candidate is to be assigned to no more than one post.
\begin{equation}
	\sum_{j=1}^{|POS|}x_{ij}\leq1,\forall i=1,2,...,|C|
\end{equation}

2. There is precisely one individual assigned to each position.
\begin{equation}
	\sum_{i=1}^{|C|}x_{ij}=1,\forall j=1,2,...,|POS|
\end{equation}

3. The team has the requisite number of members for each necessary skill. 
\begin{equation}
	\sum_{c_{i}\in C(s_{k})}x_{i j}\geq|n(s_{k})|,\forall s_{k}\in T
\end{equation}

4. The set of skills possessed by all chosen candidates is not less than the skills required by the team.
\begin{equation}
	\sum_{c_i\in C}\sum_{m\in TS(c_i)}S_m\cdot x_{ij}\geq|T|
\end{equation}

5. The selected candidates meet the skill requirements of the corresponding positions.
\begin{equation}
	\left(1-x_{ij}\right) \cdot RS\left(pos_j\right) \leq S(c_i), \forall i \in C,j \in POS
\end{equation}

6. The range of values of the decision variable.
\begin{equation}
	x_{ij}\in\{0,1\},\forall i=1,2,...,|C|,j=1,2,...,|POS|
\end{equation}

\section{The Proposed Method}
This section introduces the overall framework of RL-GP and the related improvement strategies.

\begin{figure*}[htp]
	\centering
	\includegraphics[width=0.55\textwidth]{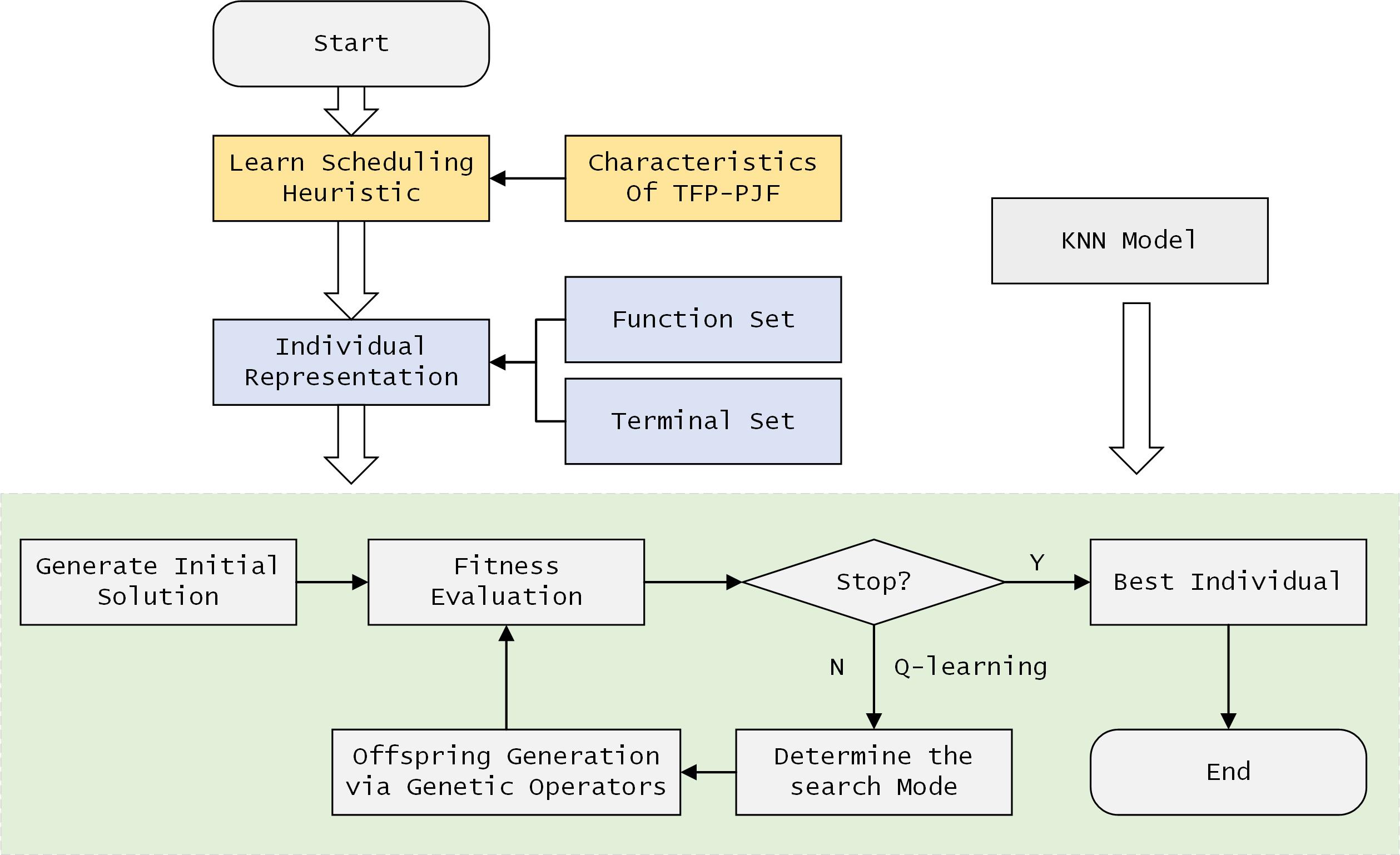}
	\caption{The flowchart of RL-GP}
	\label{The flowchart of RL-GP}
\end{figure*}

\subsection{Framework}
The TFP-PJM problem has a large number of candidate and job characteristics, which is difficult to describe them simultaneously using a simple rule. In contrast, GP can find an optimal solution by combining different rules by individuals in the population and evolving iteratively \cite{mei2022explainable}. In addition, to break the limitation of a single search pattern in GP, ensemble population strategies are implemented, with reinforcement learning employed to select search patterns from the ensemble population strategies to support the population search. Furthermore, the surrogate model is used in the individual fitness evaluation phase to improve the learning speed of RL-GP.

The flowchart of the proposed algorithm is shown in Figure \ref{The flowchart of RL-GP}. We analyze the TFP-PJM problem features and select some low-level heuristic rules, which serve as the terminal set of the problem, and select some basic operations that conform to constitute the function set. The representation of individuals in the GP will combine the terminal set and the function set into different priority functions, which are used to rank candidates, resulting in the corresponding team formation schemes. After the initialization of the algorithm, the genetic programming framework is used to iterate continuously in search of an ideal team formation solution. In the RL-GP population evolution process, the algorithm adopts ensemble population strategies and the reinforcement learning method. Specifically, four population search modes are set in the algorithm. They are respectively the population of size $N$ (denoted as $P_1$), the single elite individual population of size $N$ (denoted as $P_2$), the double elite individuals' population of size 2 (denoted as $P_3$), and the population with size $N+2$ composed of $P_1$ and $P_3$ (denoted as $P_4$). The above four search modes are used as actions and are selected by the agent before each generation of population iteration search. The reinforcement learning method involves the use of the Q-table to record the state of the agent, and the $\varepsilon$-greedy strategy for action selection \cite{kallestad2023general}. The $\varepsilon$-greedy strategy selects the action with the largest Q-value or selects a random kind of action at a particular moment by comparing random numbers and the parameter. This strategy fully exploits search information while simultaneously exploring new actions to enhance the algorithm's exploration performance.

After determining the search mode, the RL-GP population is evolved through selection, crossover, and mutation operations to generate the offspring population. The newly derived population evaluates the fitness function values using the surrogate model, and subsequently, the Q-table will be updated based on their optimization performance. To facilitate a prompt assessment of individual fitness values, the K-nearest neighbors (KNN) model is implemented as the surrogate model \cite{hildebrandt2015using}. After searching a certain number of generations, the algorithm will select the best-performing individual and determine the team formation scheme generated by it as the final scheme.

After introducing the algorithm framework of RL-GP, some methods and strategies need further explanation. The first part of this section introduces the individual representation in RL-GP and its corresponding process of encoding and decoding. The reinforcement learning method will be given in the second part. Subsequently, the ensemble population strategies will be described in detail in the third part. The fourth part elaborates on the construction of the surrogate model, while the final part introduces the learning process of RL-GP.

\subsection{Individual Representation}
Unlike the encoding of other evolutionary algorithms, GP encodes the chromosomes of individuals with the terminal set and the function set, which comprise values and operators, respectively \cite{espejo2009survey}. The terminal set consists of the relevant state features of the TFP-PJM problem, as shown in Table \ref{Terminal Set Table}. Our RL-GP uses a classical tree structure. The function set's operators serve as non-terminal nodes of the tree. The individual representation process is to combine several state features of the TFP-PJM problem in a certain order to obtain a priority function. The priority function allows all candidates to be ranked and selected one by one in order to form the project team. To select suitable positions for the candidates, the heuristic approach will select the highest matching positions among the available positions.

\begin{figure}[htp]
	\centering
	\includegraphics[width=0.15\textwidth]{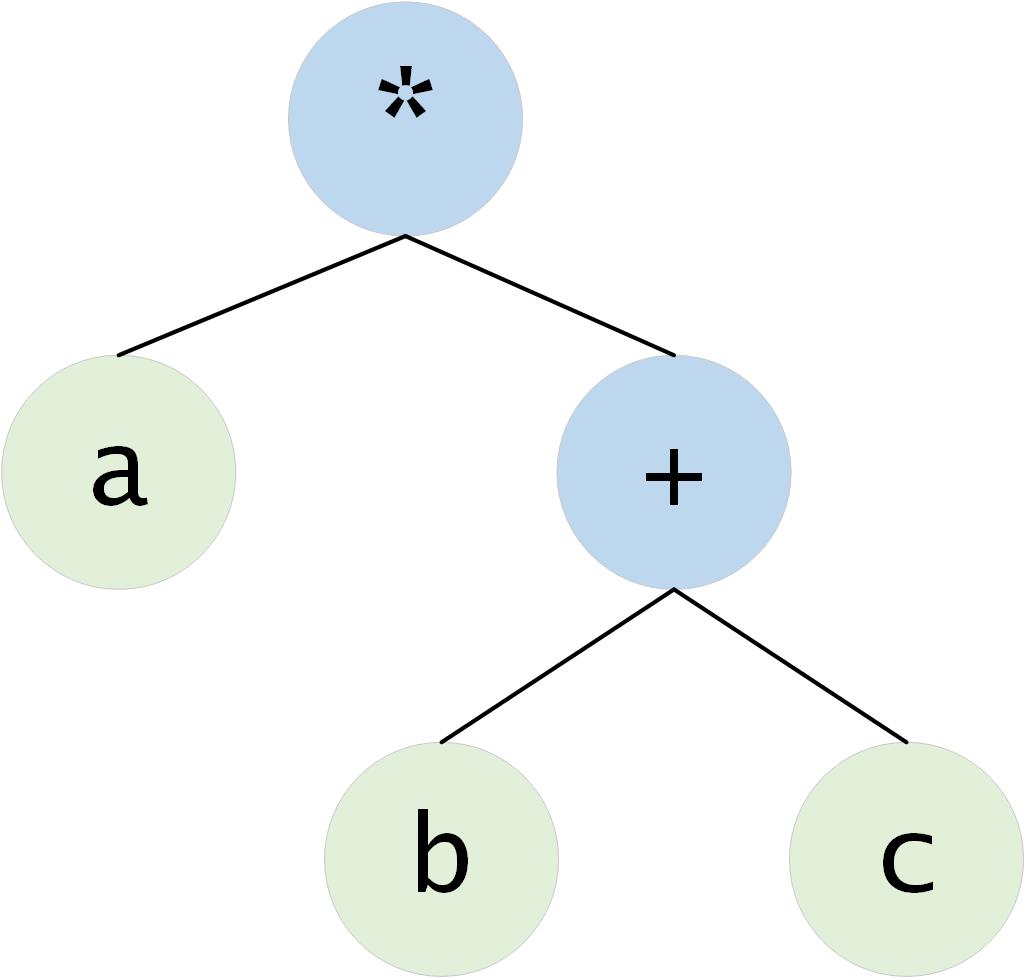}
	\caption{An example of an individual structure in RL-GP}
	\label{An example of an individual structure in RL-GP}
\end{figure}

\begin{table*}[ht]
	\small
	\caption{Terminal Set Table}
	\label{Terminal Set Table}
	\centering
	\begin{tabular}{ll}
		\toprule[1.5pt]
		Notation & Description                                                   \\
		\midrule[1pt]
		NPT      & The   number of posts set by the team                         \\
		RNP      & Remaining   number of posts                                   \\
		SC       & The   number of skills within a candidate                     \\
		SCN      & The   number of skills within a candidate that the team needs \\
		ANS      & The   average number of skills of the candidates              \\
		NSR      & The   number of skills required by the team                   \\
		NCP & The number of candidates who possess a skill required by the team        \\
		NCR      & The   number of candidates for a required skill by the team   \\
		SCW & The sum   of the scores for communication willingness of existing team members \\
		SMP      & The sum   of the scores for the arranged person-job match \\
		WEC      & The   weight of a post's evaluation of a competency indicator \\
		RAND     & A random number between   0 and 1  \\
		\bottomrule[1.5pt]                          
	\end{tabular}
\end{table*}

Figure \ref{An example of an individual structure in RL-GP} shows an example of an individual structure in RL-GP. It is a priority function $\text{a}*(\text{b}+\text{c})$ consisting of three terminals(e.g., a, b, and c) and two functions(e.g., * and +). Substituting the corresponding feature attributes of all candidates into the priority function will obtain the respective priority value. By ranking, the candidate with the highest priority value is selected to join the team.

In the process of decoding, the characteristic attributes of each candidate are substituted into the individual combination rule to get the ranking of different candidates, and the candidate with the highest priority is selected. After that, the match scores of this candidate and all positions are calculated using intuitionistic fuzzy numbers. Based on the calculation results, the candidate will be placed in the position with the highest match score. After arranging the first member, the low-level heuristic rule within the individual is modified, resulting in a new candidate priority ranking, and the next member is selected from the remaining candidates according to the updated ranking. The newly selected candidate is then placed in the position with the highest match score among the available positions. The prioritization and job selection process is repeated until all requirements for team formation are satisfied. In the individual fitness evaluation phase, RL-GP will use the surrogate model to estimate the individuals' fitness values and update the agent's status value based on the reward and search performance. The following parts will describe the ensemble of different population search strategies in reinforcement learning, as well as the specific process of using reinforcement learning to guide the evolution of genetic programming populations.

\subsection{Reinforcement Learning Method}

\label{Reinforcement Learning Method}

In the proposed algorithm, we use reinforcement learning methods to improve the efficiency of algorithmic population search by selecting suitable search modes from ensemble population strategies. Reinforcement learning involves a learning process in which an agent interacts with its environment, obtaining feedback signals and rewards to optimize its decision-making process and maximize the cumulative reward \cite{dabbaghjamanesh2020reinforcement}. The proposed algorithm employs a simple structured Q-learning method in reinforcement learning, where the agent's state is recorded in the Q-table. Population search is executed based on the selected search mode, and the agent's reward is calculated based on the new rules for team formation discovered by the search and recorded in the Q-table to guide subsequent decisions. When selecting the search pattern, the $\varepsilon$-greedy strategy is introduced, augmenting the algorithm's exploration capability by utilizing existing information while continuing to explore untried actions.

\begin{figure*}[htp]
	\centering
	\includegraphics[width=0.55\textwidth]{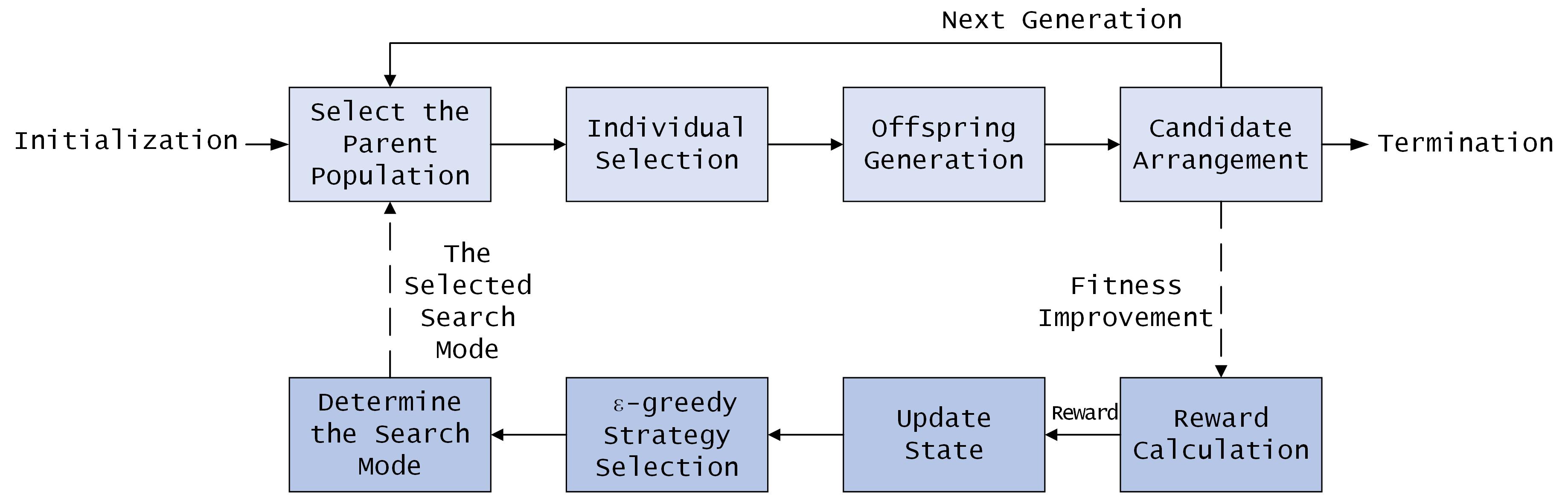}
	\caption{Dataflow between the Q-learning and GP}
	\label{Dataflow between the Q-learning and GP}
\end{figure*}

Figure \ref{Dataflow between the Q-learning and GP} reflects the principle of RL-GP population evolution in collaboration with Q-learning, including two critical processes. Reinforcement learning assists the RL-GP population to determine the search mode, which can make the population more focused on possible optimal solutions and thereby improve search efficiency and quality. At the same time, the evolutionary results of the GP population are fed back to the agent to update the Q-table, further directing subsequent search decisions for superior search results \cite{da2018parallel}. The two processes continuously interact in a cycle to enable the algorithm to find the optimal rules and get high-quality team formation rules. In this process, the parent population evolves with a specific search mode and generates the offspring population through evolutionary operators such as crossover and mutation. The best individual in the offspring population outputs a new solution, corresponding to the best fitness value of the contemporary generation. The agent compares the best individual fitness values of two adjacent generations, calculates the reward for choosing a search mode in this state, and updates the Q-table. As it keeps trying and updating, the agent makes optimal decisions through the Q-table combined with the $\varepsilon$-greedy strategy to guide the population evolution more efficiently. The results of each evolutionary generation are evaluated and the search strategy is further optimized until an approximate optimal solution is found or a predetermined number of iterations is reached. This interactive algorithm is designed to utilize the information obtained from the population search to find an optimal strategy.

The RL-GP population search is only related to the current state and satisfies the conditions for constructing Markov Decision Processes (MDP) \cite{doltsinis2014mdp}. An MDP consists of four components: state, action, reward, and value function. The RL-GP algorithm associates the state of the agent with the improvement of the best individual fitness value for each generation of the population. This association can be categorized into two types: those with improvement and those without improvement. The action setting in RL-GP contains four population search modes, which are integrated into the population strategy. After each generation of population evolution, the best-performing individual is selected, and the reward is calculated by the difference in individual fitness values at different time points. This calculation of $R_t$ adheres to the following relationship:
\begin{equation}
	R_t=f_t^*(S_t,A_t)-f_{t-1}^*(S_{t-1},A_{t-1})
	\label{reward}
\end{equation}
where $f_t^*(S_t, A_t)$ represents the fitness value of the best individual obtained by population evolution after the agent chooses the search mode $A$ for RL-GP population in the state $S$ at the time point $t$. The difference between it and the best individual fitness value at the time point $t-1$ is the immediate reward $R_t$ at the time point $t$.

During the population search process, the agent selects different search modes based on the current state and calculates the reward according to the search performance. The agent stores the calculation results in the Q-table and makes the next action selection according to the new state. The Q-value is updated by the following formula:
\begin{equation}
	\begin{split}
		Q_{t+1}(S_t,A_t)=Q_t(S_t,A_t)+\alpha[R_t+\\ \gamma\max\limits_{a\in A}Q(S_{t+1},a)-Q(S_t,A_t)]
	\end{split}
	\label{q-value}
\end{equation}

\begin{figure*}[htp]
	\centering
	\includegraphics[width=0.55\textwidth]{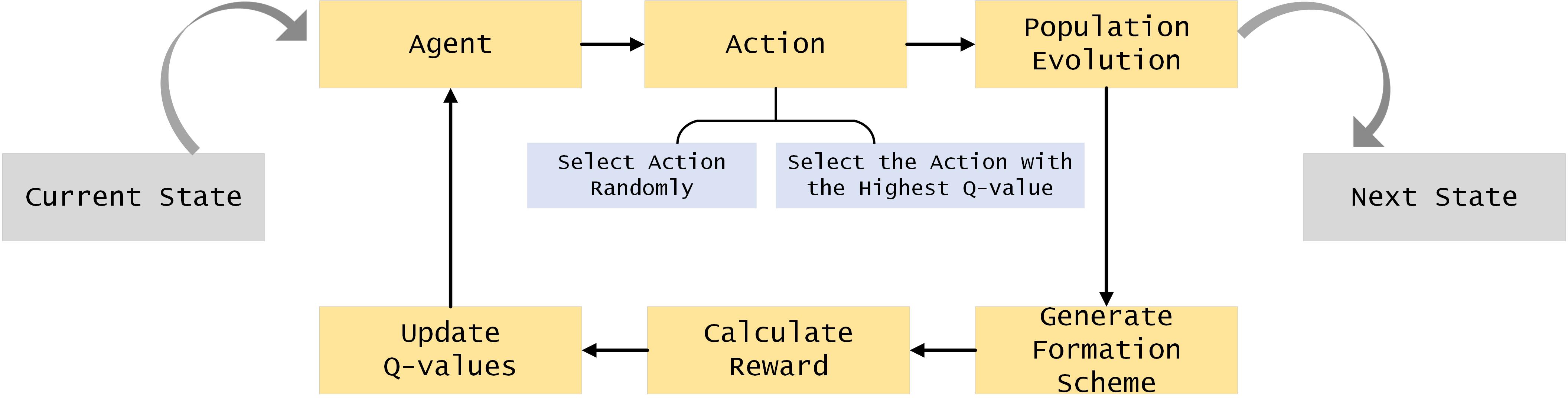}
	\caption{Decision-making Process of Q-learning Method}
	\label{Q-learning Method Decision-making Process}
\end{figure*}

There are two adjustable parameters in the function, $\alpha$ represents the learning rate and $\gamma$ represents the discount factor. The updating process is a dynamic planning method based on the Bellman equation, known as the Bellman Equation \cite{moon2020generalized}. This equation provides the optimal decision value based on the current state, which is equivalent to the expected value of the optimal decision value in the subsequent state, added to the immediate reward obtained in the current state. As can be seen from Figure \ref{Q-learning Method Decision-making Process}, the next state is determined after the RL-GP population in the current state selects a specific search mode. The $\max\limits_{a\in A}Q(S_{t+1}, a)$ in the formula refers to the maximum Q value attainable in all actions in the next state.

In the RL-GP algorithm, the agent realizes a $\varepsilon$-greedy strategy for action selection and picks the appropriate search mode. This strategy involves setting a constant threshold $\varepsilon$ that takes values from 0 to 1 to control the degree of randomness. The $\varepsilon$-greedy strategy is employed when $\varepsilon$ equals 0, which means that the action is selected exactly according to the currently known optimal strategy, called the $\varepsilon$-greedy strategy. Conversely, when $\varepsilon$ is 1, it represents a completely random choice of action, which is called a completely random strategy. For other values of $\varepsilon$, the agent generates a random number from 0 to 1 before each decision. When the random number is less than $\varepsilon$, the agent will randomly select one of all population search strategies; conversely, the population strategy with the highest Q-value will be selected for the search. The pseudo-code for the $\varepsilon$-greedy strategy is shown in Algorithm 1.

\begin{algorithm}[ht]
	\label{varepsilon-greedy Strategy}
	\caption{$\varepsilon$-greedy Strategy}
	\LinesNumbered
	\KwIn{state ${S_{t}} $, action set $A$, Q-table $Q$, parameter $\varepsilon$. }
	\KwOut{action $a$}
	$rand \leftarrow $Generate a random number between 0 to 1\;
	\eIf{$\varepsilon \le rand$ }{
		$a \leftarrow $Select an action randomly from the action set $A$\;
	}{
		$a \leftarrow $Select the action with the largest Q-value from Q-table $Q$ based on the state ${S_{t}} $\;
	}
\end{algorithm}

\subsection{Ensemble Population Strategy}

\label{Ensemble Population Strategy}

Population search is a widely used global optimization method that includes various meta-heuristic and hyper-heuristic searches, including genetic programming. Compared with traditional individual search, it can reduce the probability of the algorithm converging to a local optimum and improve the search efficiency of the algorithm through initializing a population of potential problem solutions, which undergoes continuous iterations and co-evolution. 

\begin{figure}[htp]
	\centering
	\includegraphics[width=0.4\textwidth]{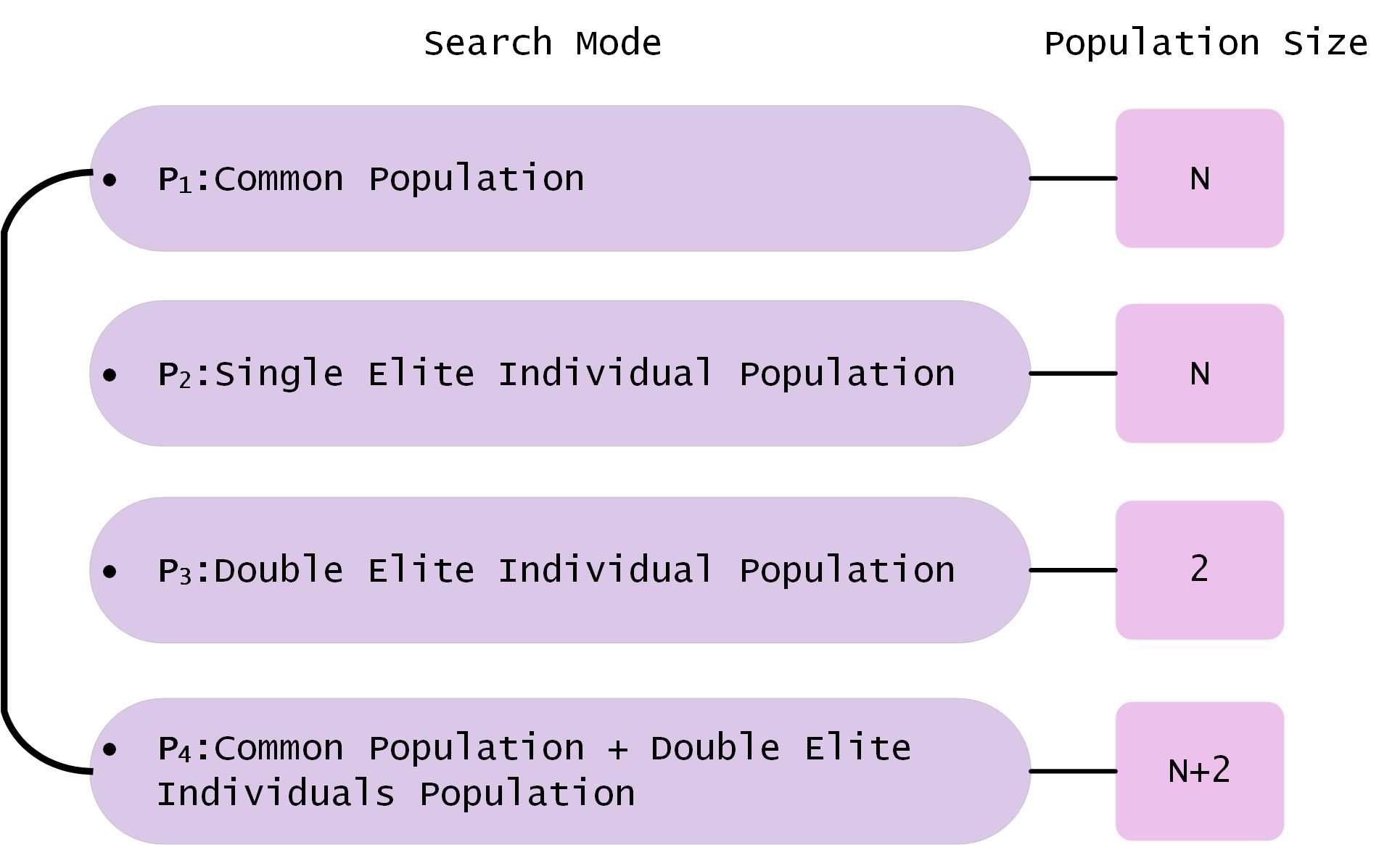}
	\caption{Four Search Modes}
	\label{Four Search Modes}
\end{figure}

However, for genetic programming, it is more difficult to find high-quality solutions when solving complex combinatorial optimization problems using only a single population search model \cite{ye2017novel}. To improve the search performance of the algorithm, an ensemble population strategy is employed in RL-GP \cite{wu2019ensemble}. Specifically, before the start of each generation of population search, the algorithm will use the reinforcement learning method to select one of the population search modes for the contemporary search. The ensemble of the four population strategies in the RL-GP algorithm is shown in Figure \ref{Four Search Modes}.

$\bullet$ $\mathbf{P_1}$: (The population size of common population is $N$.) This is the most common form of population evolution approach used in the genetic programming algorithm for initialization. This approach generates an offspring population through evolutionary operations such as selection, crossover, and mutation.

$\bullet$ $\mathbf{P_2}$: (The population size of single elite individual population is $N$.) To expedite the convergence of RL-GP, we use the global best individual to randomly replace an individual in $P_1$ to form $P_2$. After each generation of population evolution, we compare the fitness of the contemporary best individual with the global elite individual. If the best individual outperforms the elite individual, we use the new elite individual to replace the elite individual in the original $P_2$ population.

$\bullet$ $\mathbf{P_3}$: (The population size of  double elite individuals population is $2$.) The $P_3$ population consists of the global best individual and the best individual from the preceding generation. This population search mode is similar to the neighborhood search and has the potential to mine high-quality solutions in the local solution space.

$\bullet$ $\mathbf{P_4}$: (The population size of common population $+$ double elite individuals population is $N+2$.) The $P_4$ population consists of $P_1$ and $P_3$ together. Such a population strikes a balance between global and local search abilities. Accordingly, this population requires increased computational resources for the algorithm search.

The four population search strategies are integrated into the genetic programming framework and selected using the reinforcement learning method, making the RL-GP algorithm highly adaptive in its search capabilities. The agent can dynamically adjust the search strategy according to the current search performance to improve the effectiveness of the search, thus increasing the chances of the RL-GP algorithm finding the optimal solution.

\subsection{Surrogate Model}

\label{Surrogate Model}

Since the complexity of the TFP-PJM problem is NP-hard, generating a team formation scheme requires complex constraints checking. The process of constraint evaluation and verification is computationally intensive and can significantly affect the learning efficiency of the algorithm \cite{zhang2021surrogate}. As the size of the problem increases, the learning time of the algorithm also increases proportionately. In view of the computationally expensive nature of the fitness function evaluation, it is essential to identify an alternative approach. The approach of surrogate modeling then provides a solution idea for fast evaluation of individual fitness functions. Refer to \cite{hildebrandt2015using}, we use the k-Nearest Neighbor (KNN) model to approximate the fitness function values. This simple machine learning model computes the average of the fitness values of the $k$ nearest individuals to a new individual by calculating the distances between individuals, thus estimating the predicted value of the new individual. Such an adaptation evaluation method has low time complexity and can speed up the iterative population search process.

Since the RL-GP algorithm uses an individual representation that differs from other population-based search algorithms, the tree structure does not allow the genotype to be directly input into the surrogate model. Therefore, refer to \cite{hildebrandt2015using}, we generate a decision vector by using the reference rule and other rules, where the reference rule entails an employee's suitability for a job, ranked in descending order, and the other rules stem from RL-GP learning. To construct the decision vector value for each job dimension, the candidates' ranking results are obtained using the reference rule and the rules obtained by RL-GP learning. After that, the candidate whose ranking is the first in the learned rule is used, and the ranking result under the reference rule is taken as the decision vector value of this dimension. After generating the decision vectors, denoted as $\mathbf{d^A}$ and $\mathbf{d^B}$, using the rules A and B obtained through RL-GP learning, the distance between the two vectors, with a number $i$ of positions, is calculated as follows:
\begin{equation}
	\begin{split}
		D(\mathbf{d^A},\mathbf{d^B})=\sqrt{\sum_{i=1}^k\left(d_i^A-d_i^B\right)^2}
	\end{split}
	\label{distance}
\end{equation}

The distance between the predicted individual and other individuals in the training set is calculated by the KNN model. $k$ individuals closest to the individual are found, and the average of the fitness values is calculated to serve as the predicted value of the individual. For the selection of $k$, it is necessary to consider the number of actual individuals in the training set, which can be modified flexibly in the operation of the model \cite{zhang2017efficient}. To ensure the prediction effectiveness and computational efficiency of the KNN model, the threshold $\mu$ is used to decide whether to update the surrogate model. Specifically, at intervals of every $\mu$ generation, RL-GP evaluates the performance of the best individuals found in each previous generation of search through the real evaluation model. These individuals, along with their corresponding real fitness function values, are subsequently appended to the training set, and the KNN model is updated. Through online learning, the accuracy of the model prediction can be continuously improved to ensure the quality of the best individual ultimately produced by the algorithm.

\subsection{Learning Process of RL-GP}

The framework and some strategies in RL-GP are introduced in the previous parts. In this part, the learning process of RL-GP is given in Algorithm 2.

\begin{algorithm}[htb]
	\label{Learning Process of the Reinforcement Learning-assisted Genetic Programming Algorithm}
	\caption{Learning Process of the RL-GP}
	\LinesNumbered
	\KwIn{ population size $N_p$, learning rate $\alpha$, discount factor $\gamma$, action set $A$, $\varepsilon$, $k$, threshold $\mu$}
	\KwOut{ $gobal\_{best}\_indi$.}
	$P \leftarrow$Initialize the population\;
	Initialize the surrogate model\;
	Initialize Q-table and set $t \leftarrow 0$\;
	Set $iter \leftarrow 0$\;
	\While{the termination criteria are not met}{
		$ a \leftarrow $ Select the search mode in \ref{Ensemble Population Strategy} using $\varepsilon$-greedy strategy in Algorithm 1\;
		$P \leftarrow$ Adjust the population composition according to the action $a$\;
		Select individual from $P$\;
		$ P' \leftarrow $ Population evolution by using crossover and mutation operators $\left(P\right)$\;
		Evaluate fitness using surrogate model in \ref{Surrogate Model}\;
		$\Omega \leftarrow $Record the best individual in the offspring generation\;
		$R_{t} \leftarrow $Calculate reward by using Eq. \eqref{reward}\;
		$Q_{t+1} \leftarrow $Calculate the Q-value and update the Q-table by using Eq. \eqref{q-value}\;
		
		\If{$local_{best} \geq gobal_{best}$}{
			$gobal_{best} \leftarrow  local_{best}$\;
			$gobal\_{best}\_indi\leftarrow local\_{best}\_indi$\;
		}
		\If{$mod\left(iter, \mu\right)$ == 0}{
			$ f' \leftarrow $ Evaluate fitness using model in \ref{Model} $\left(\Omega\right)$\;
			Update the surrogate model $\left(f'\right)$\;
			$\Omega \leftarrow \left[ \ \right]$\;
			$iter \leftarrow 0$\;
		}
		$t \leftarrow t+1$\;
		$iter \leftarrow iter+1$
	}
\end{algorithm}

As shown in Algorithm 2, after RL-GP completes the individual fitness evaluation using the surrogate model, the highest-performing individuals within the population are recorded (Line 9). These recorded individuals are subsequently utilized to calculate the real fitness function (Line 16) in order to update the surrogate model (Line 17).

\section{Experiment}

\begin{figure*}[htbp]
	\centering
	\subfloat[Boxplot of the learning time for instances with 25 positions]{\includegraphics[width=.3\textwidth]{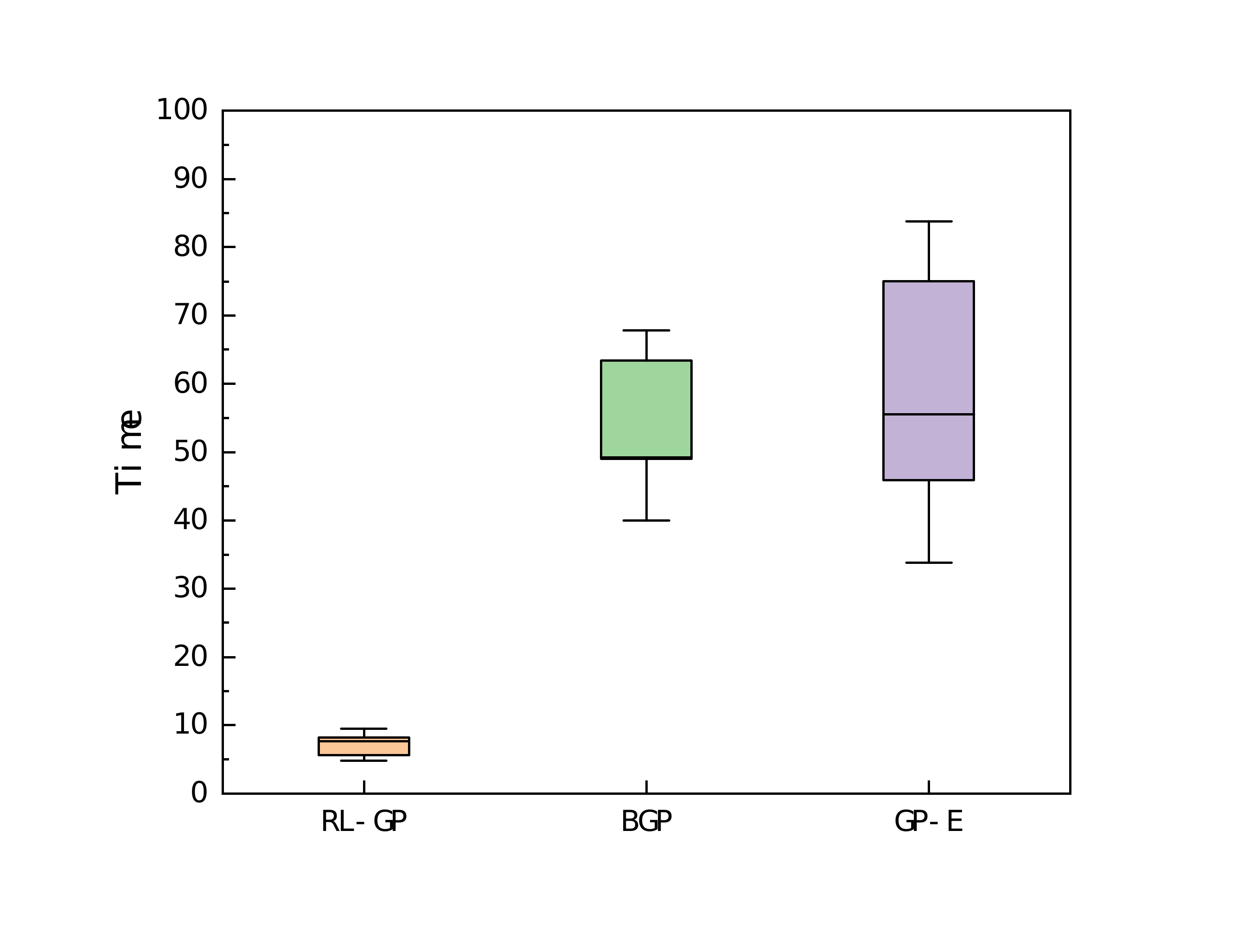}} \qquad
	\subfloat[Boxplot of the learning time for instances with 50 positions]{\includegraphics[width=.3\textwidth]{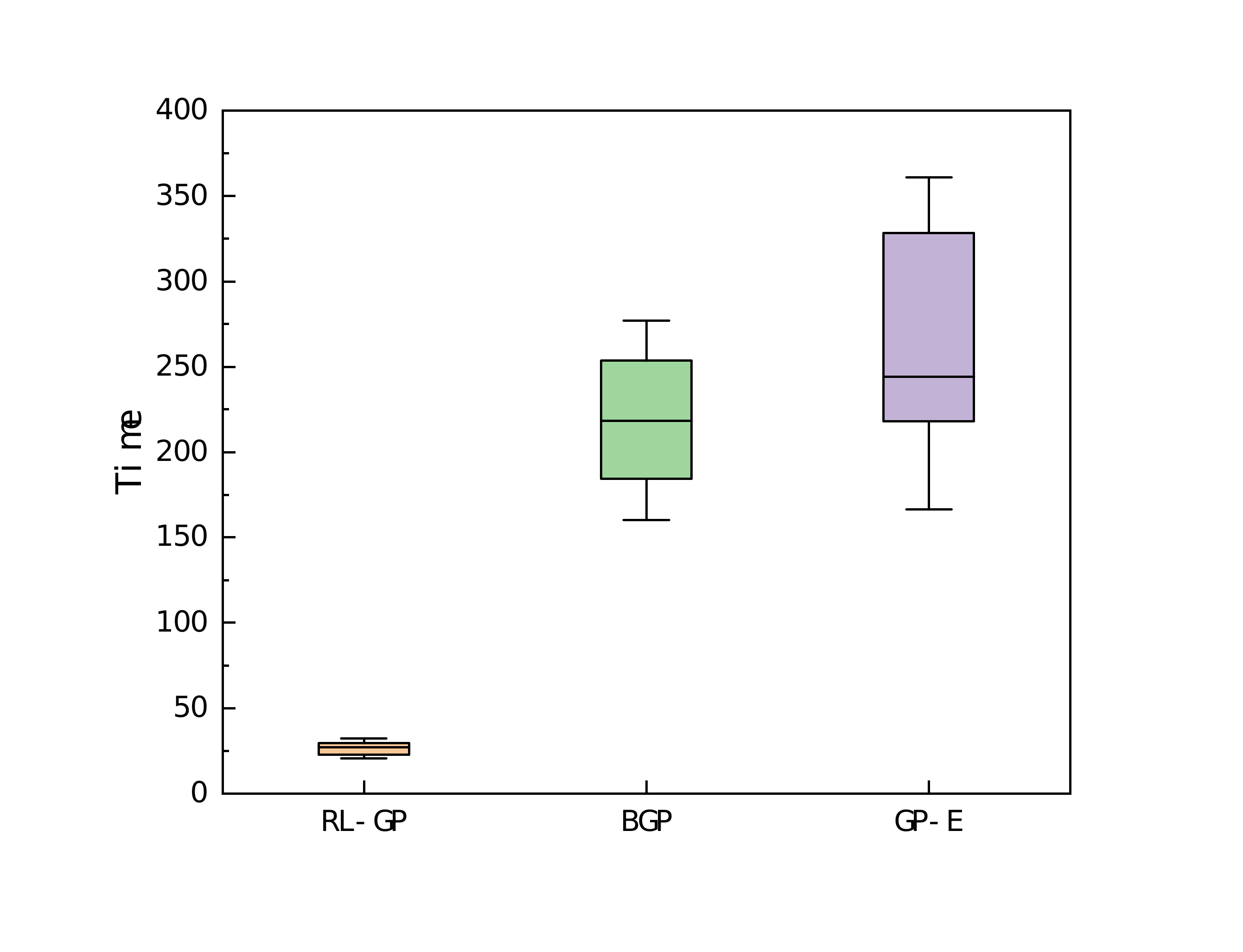}} \qquad
	\subfloat[Boxplot of the learning time for instances with 75 positions]{\includegraphics[width=.3\textwidth]{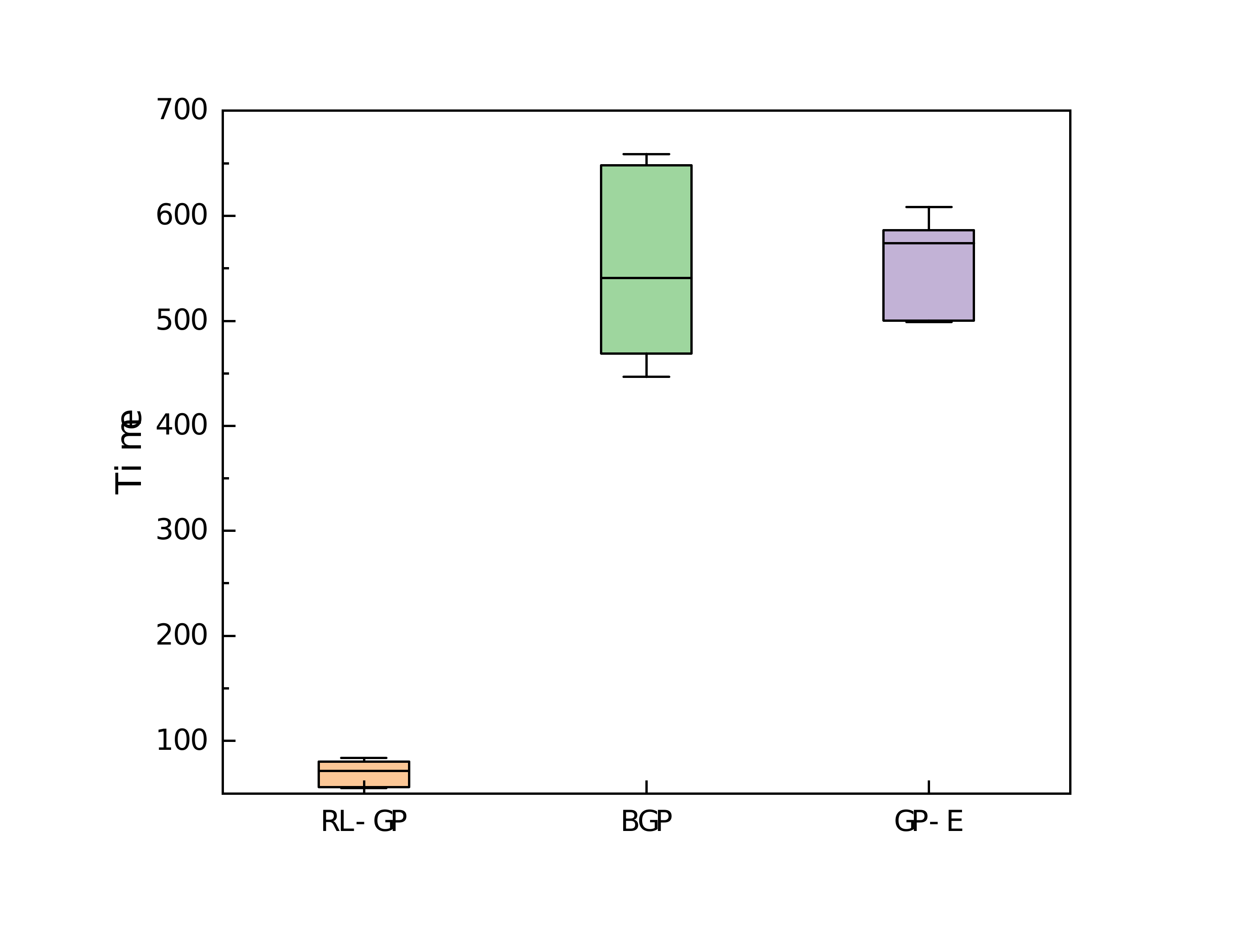}} \qquad
	\subfloat[Boxplot of the learning time for instances with 100 positions]{\includegraphics[width=.3\textwidth]{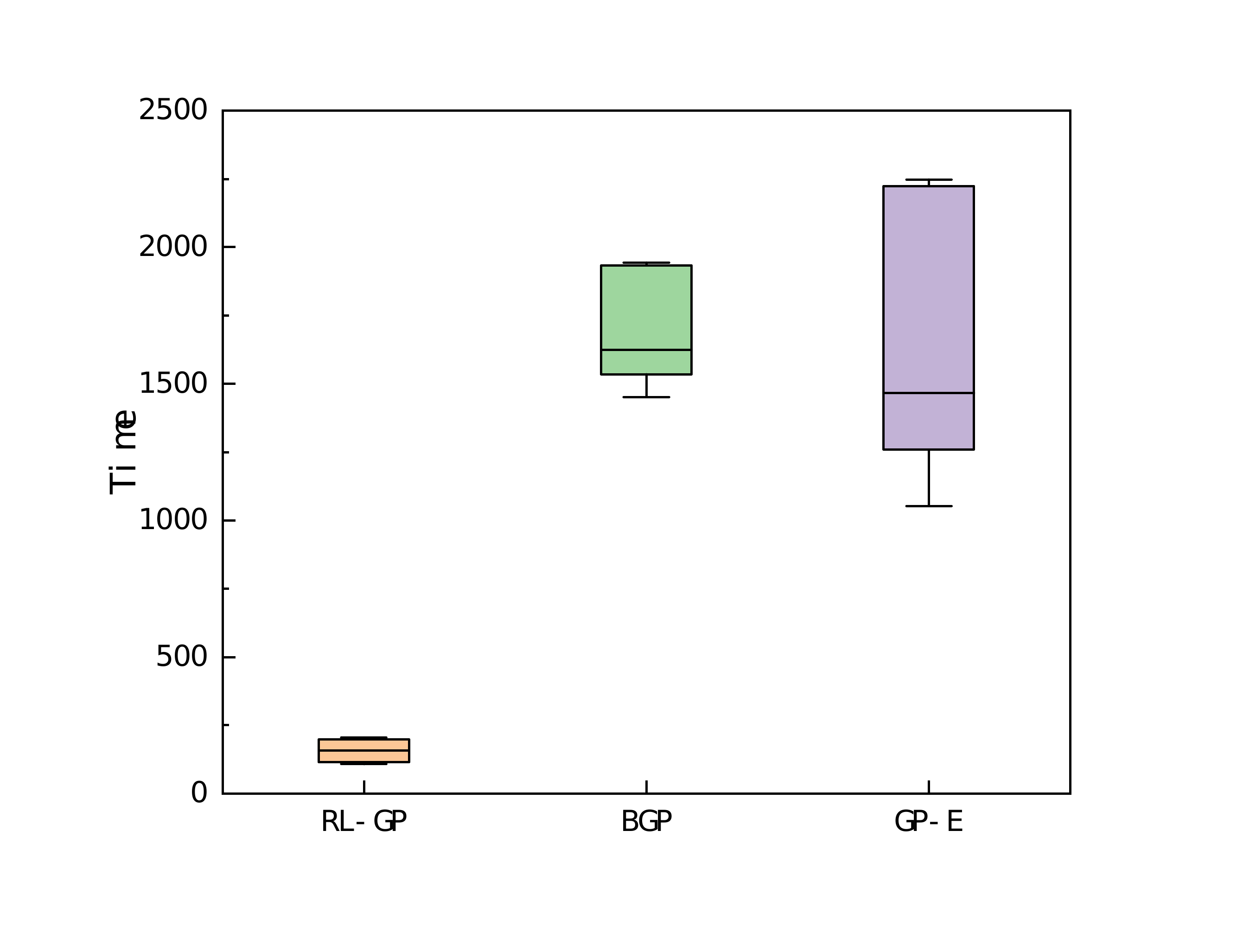}} \qquad
	\subfloat[Boxplot of the learning time for instances with 125 positions]{\includegraphics[width=.3\textwidth]{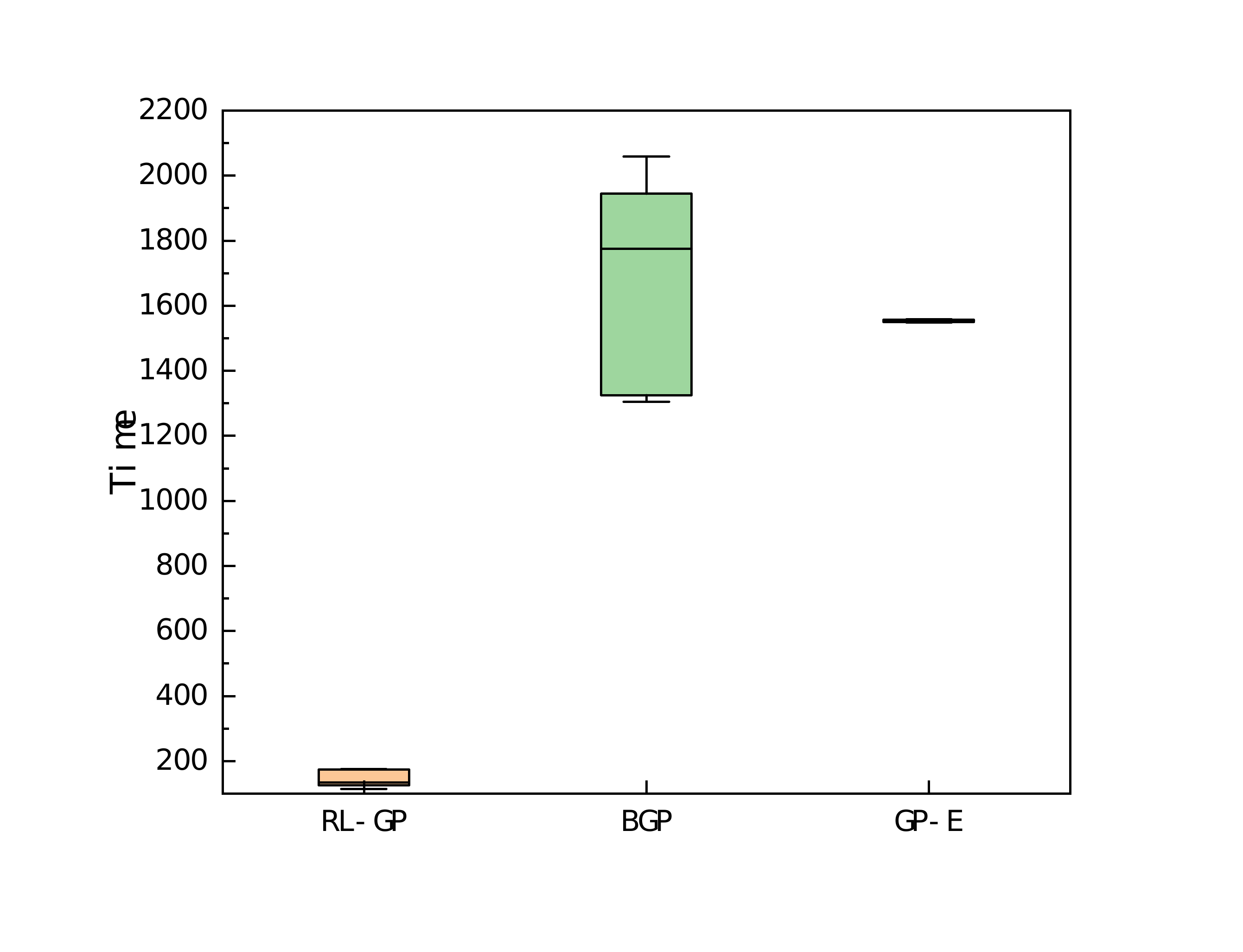}} \qquad
	\subfloat[Boxplot of the learning time for instances with 150 positions]{\includegraphics[width=.3\textwidth]{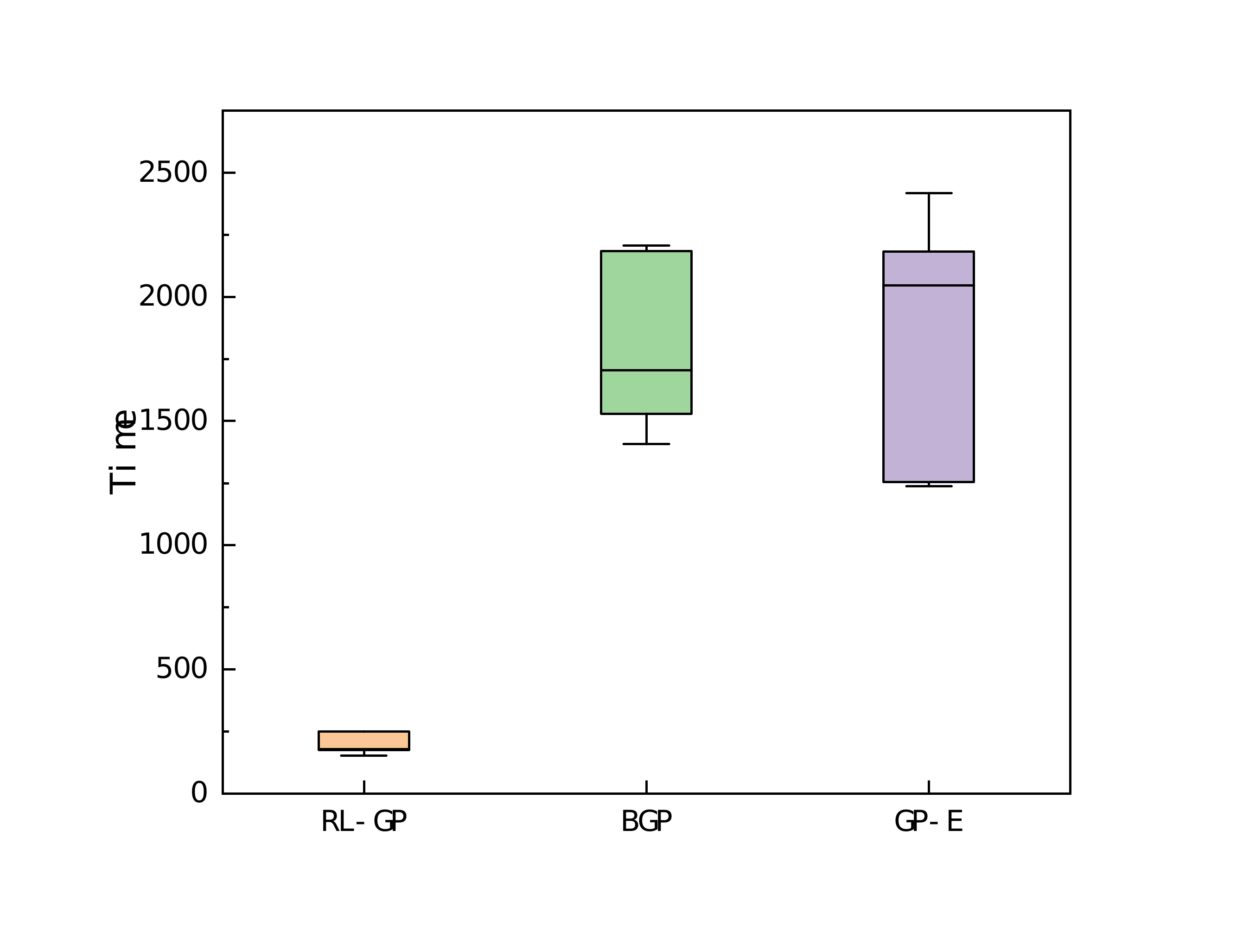}} \qquad
	\subfloat[Boxplot of the learning time for instances with 175 positions]{\includegraphics[width=.3\textwidth]{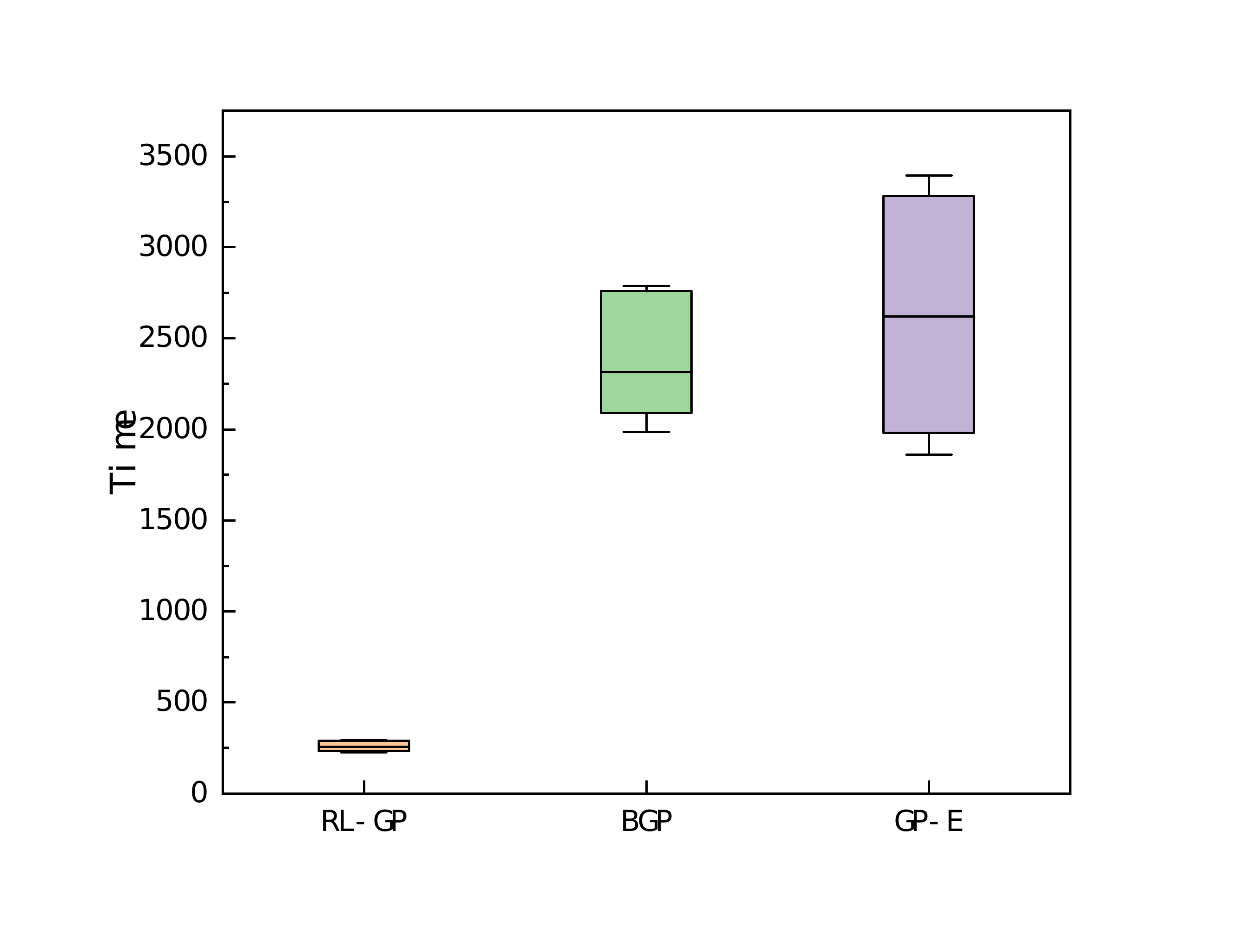}} \qquad
	\subfloat[Boxplot of the learning time for instances with 200 positions]{\includegraphics[width=.3\textwidth]{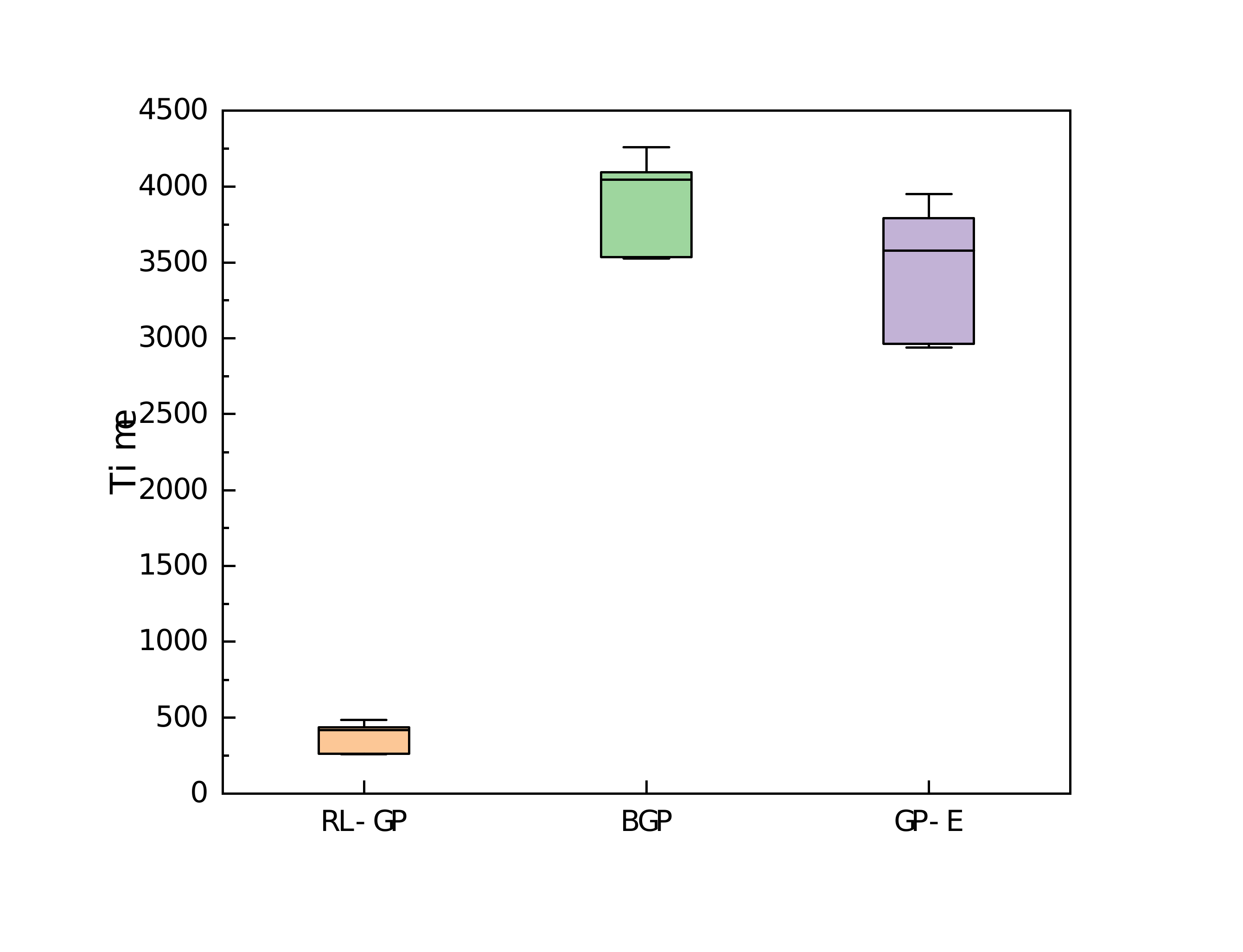}} \qquad
	\subfloat[Boxplot of the learning time for instances with 225 positions]{\includegraphics[width=.3\textwidth]{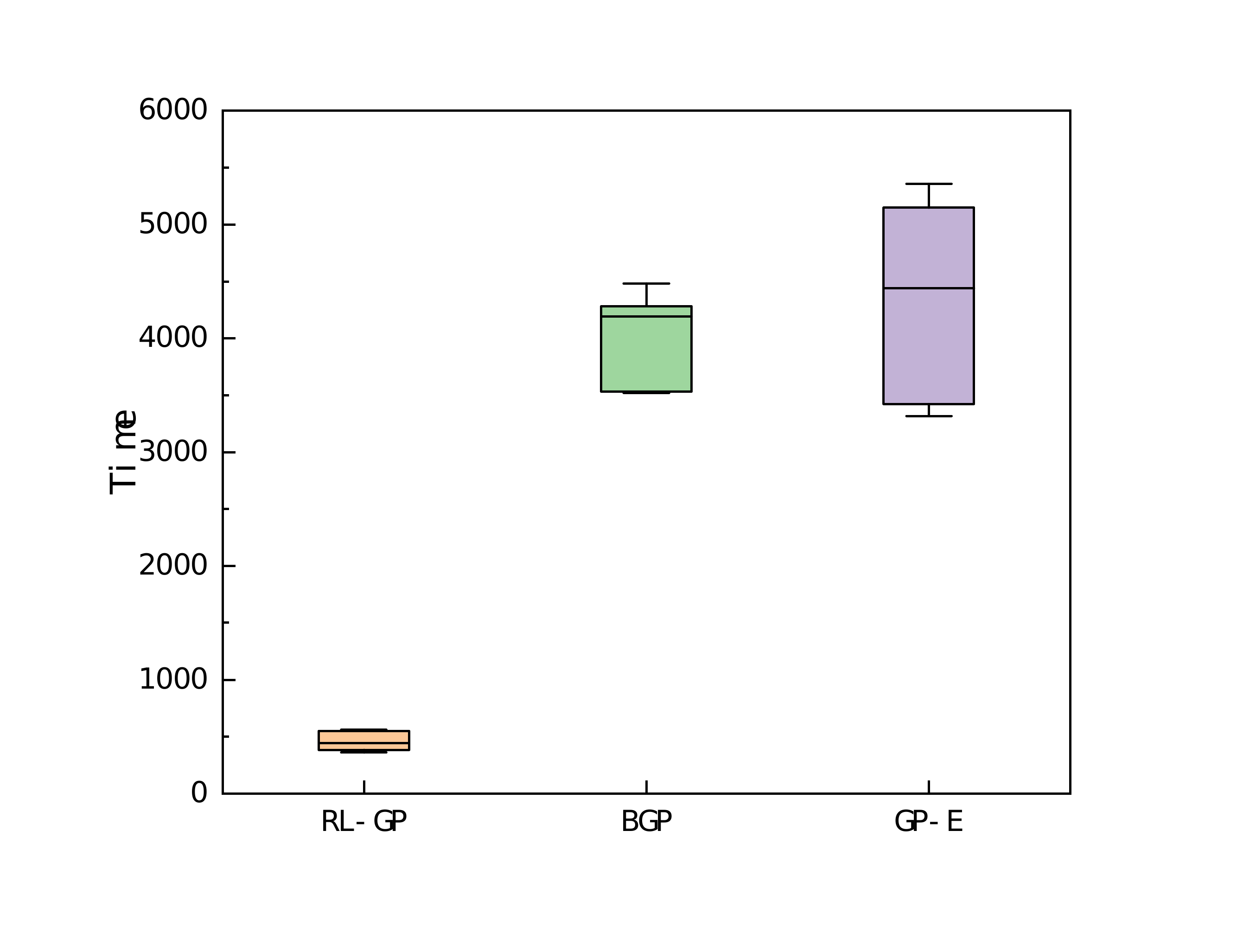}} 
	\caption{Boxplots of the learning time for instances with different scale positions}
	\label{Boxplots of the learning time for instances with different scale positions}
\end{figure*}
In this section, the experimental setup, experimental results, and discussion will be presented.

\subsection{Experimental Setup}
Experimental environment: In this section, a series of experiments are used to verify the effectiveness of the RL-GP algorithm. All algorithms are performed on a Core(TM) i5-8265U CPU 1.60GHz, Windows 10 OS computer with Python 3.11 code environment.

Experimental scenarios: There is no public benchmark for the TFP-PJM problem, so we use a random generation approach to obtain scenarios of different scales. To facilitate the representation of diverse instances, we adopt the form of “P-ID” to represent instances, where “P” represents the number of jobs and “ID” represents the internal number of the same job scale.

Comparison algorithms: To verify the solution performance of the RL-GP algorithm, various GP algorithms, construction heuristics, and search algorithms are used in the experiments. We choose traditional GP (denoted as BGP), GP using an external repository (denoted as GP-E), two constructive heuristics algorithms (denoted as CH1 and CH2 respectively), improved genetic algorithm (IGA) \cite{wang2022improved}, and variable neighborhood search algorithm (VNS) \cite{hansen2019variable} as comparison algorithms. The constructive heuristics algorithms CH1 and CH2 are employed to generate preferences for candidate selection based on different criteria. CH1 generates a preference order for selecting candidates, ranked from highest to lowest in terms of the number of skills candidates meet on the team. CH2 generated a preference order for selecting candidates based on the person-job match score, also ranked in descending order.

Algorithm parameter settings: After cross-validation experiments of the parameters, we make the following settings. In the RL-GP, the population size is set to 100, the maximum number of iterations is set to 100, the initialization method is ramped-half-and-half, the crossover is the two-point crossover, the mutation is the single-point mutation, the crossover probability is set to 0.9, the mutation probability is set to 0.1, the discount factor is set to 0.9, and the learning rate is set to 0.01, $k$ is set to 5. Function set contains $\{+,-,\ast,/,\sin,\cos,\max,\min\}$. It is worth noting that the division operation returns a result of 1 using the protection policy when the denominator is 0. The other GP algorithms’ parameters are set the same as RL-GP, and the search algorithms’ parameters are set consistent with the related literature.

\begin{table*}[ht]
	\small
	\centering
	\caption{Comparison results of RL-GP with BGP and GP-E}
	\label{Comparison results of RL-GP with BGP and GP-E}
	\begin{tabular}{llllll}
		\toprule[1.5pt]
		\multicolumn{1}{c}{\multirow{2}{*}{Instance}} & \multicolumn{1}{c}{\multirow{2}{*}{RL-GP}} & BGP &  & GP-E &  \\
		\cmidrule(l){3-4} \cmidrule(l){5-6}
		\multicolumn{1}{c}{} & \multicolumn{1}{c}{} & TE               & Gap (\%)     & TE               & Gap (\%)     \\
		\midrule[1pt]
		25-1  & \textbf{14.50}  & \textbf{14.50} & 0.00  & \textbf{14.50} & 0.00  \\
		25-2  & \textbf{14.75}  & \textbf{14.75} & 0.00  & \textbf{14.75} & 0.00  \\
		25-3  & \textbf{14.50}  & 13.86          & -4.43 & 13.86          & -4.43 \\
		50-1  & \textbf{25.02}  & 24.88          & -0.56 & \textbf{25.02} & 0.00  \\
		50-2  & \textbf{24.65}  & 24.64          & -0.05 & 24.52          & -0.56 \\
		50-3  & \textbf{25.02}  & 25.04          & 0.11  & 24.93          & -0.36 \\
		75-1  & \textbf{37.66}  & 36.64          & -2.72 & 36.73          & -2.49 \\
		75-2  & \textbf{37.66}  & 36.32          & -3.56 & 36.53          & -3.02 \\
		75-3  & \textbf{36.86}  & 35.88          & -2.68 & 35.88          & -2.68 \\
		100-1 & \textbf{47.55}  & \textbf{47.55} & 0.00  & \textbf{47.55} & 0.00  \\
		100-2 & 47.48           & 46.89          & -1.23 & \textbf{47.79} & 0.66  \\
		100-3 & \textbf{47.51}  & 47.36          & -0.31 & 46.61          & -1.89 \\
		125-1 & \textbf{59.16}  & 57.68          & -2.50 & 59.08          & -0.15 \\
		125-2 & \textbf{58.68}  & 57.41          & -2.17 & 58.64          & -0.07 \\
		125-3 & \textbf{59.16}  & 58.97          & -0.33 & 58.57          & -1.00 \\
		150-1 & \textbf{69.69}  & 67.14          & -3.66 & 69.66          & -0.04 \\
		150-2 & \textbf{69.38}  & 67.41          & -2.83 & 68.43          & -1.36 \\
		150-3 & \textbf{68.68}  & 68.57          & -0.17 & 67.95          & -1.07 \\
		175-1 & \textbf{80.38}  & 79.98          & -0.49 & 79.99          & -0.48 \\
		175-2 & \textbf{78.88}  & 78.57          & -0.39 & 78.57          & -0.39 \\
		175-3 & \textbf{79.86}  & 78.56          & -1.63 & 79.44          & -0.53 \\
		200-1 & \textbf{91.18}  & \textbf{91.18} & 0.00  & \textbf{91.18} & 0.00  \\
		200-2 & \textbf{90.52}  & 89.78          & -0.81 & 88.72          & -1.98 \\
		200-3 & \textbf{90.42}  & 88.76          & -1.84 & 89.96          & -0.51 \\
		225-1 & \textbf{102.42} & 98.62          & -3.72 & 98.44          & -3.89 \\
		225-2 & \textbf{100.66} & 99.39          & -1.26 & 99.97          & -0.69 \\
		225-3 & \textbf{100.94} & 100.55         & -0.39 & 100.93         & -0.02\\
		\bottomrule[1.5pt]
	\end{tabular}
\end{table*}

\begin{figure*}[htbp]
	\centering
	\subfloat[Results for instances with 25 positions]{\includegraphics[width=.3\textwidth]{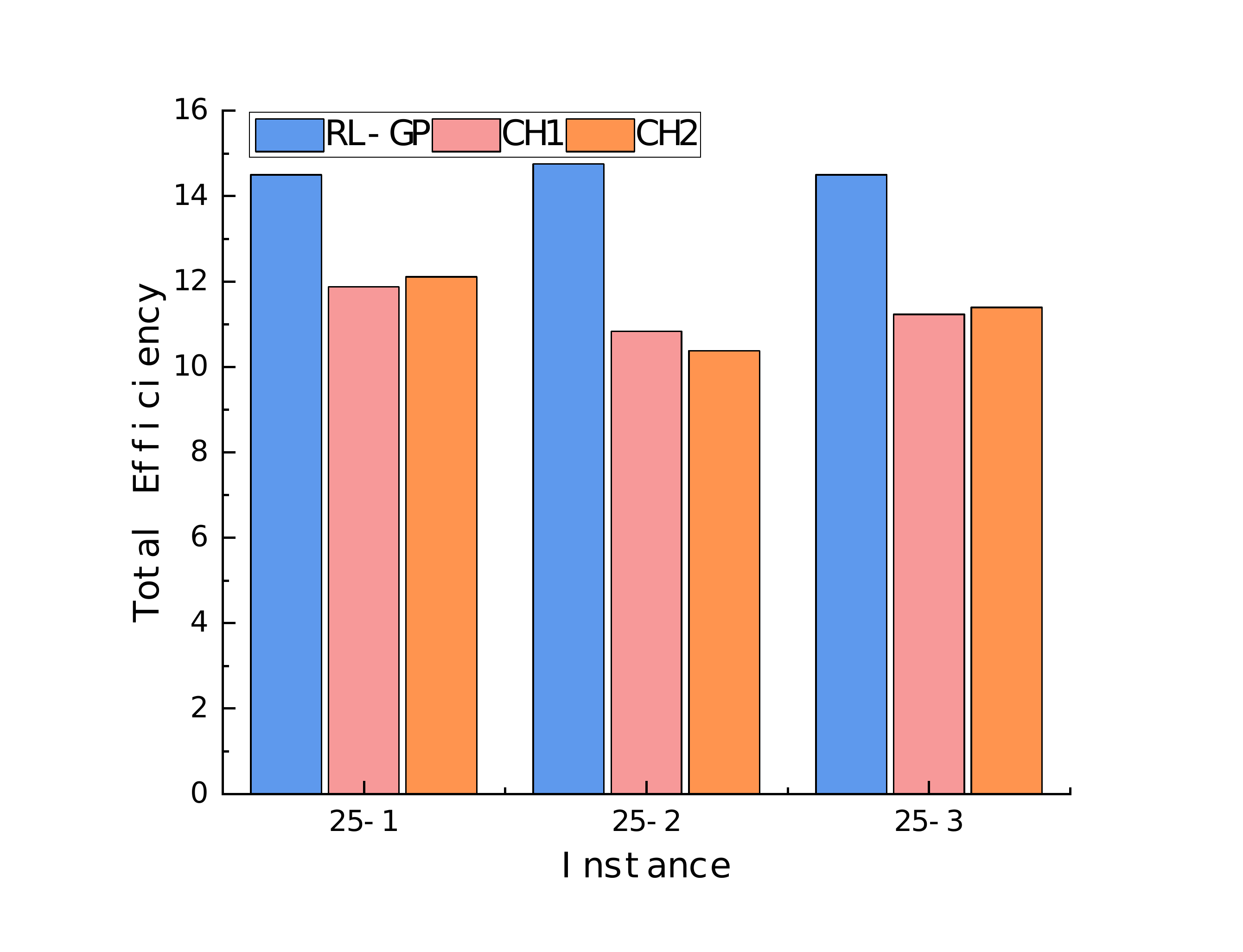}} \qquad
	\subfloat[Results for instances with 50 positions]{\includegraphics[width=.3\textwidth]{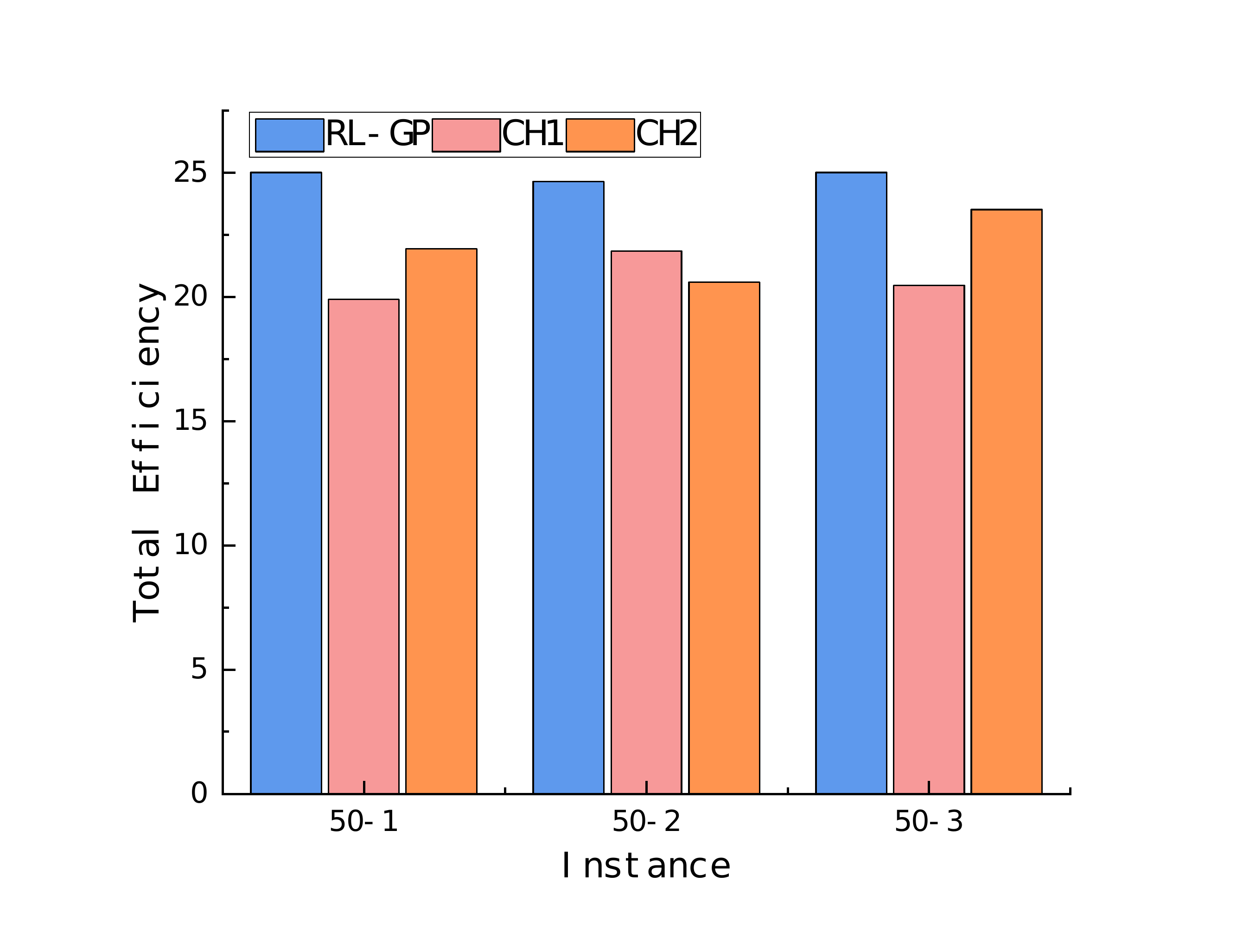}} \qquad
	\subfloat[Results for instances with 75 positions]{\includegraphics[width=.3\textwidth]{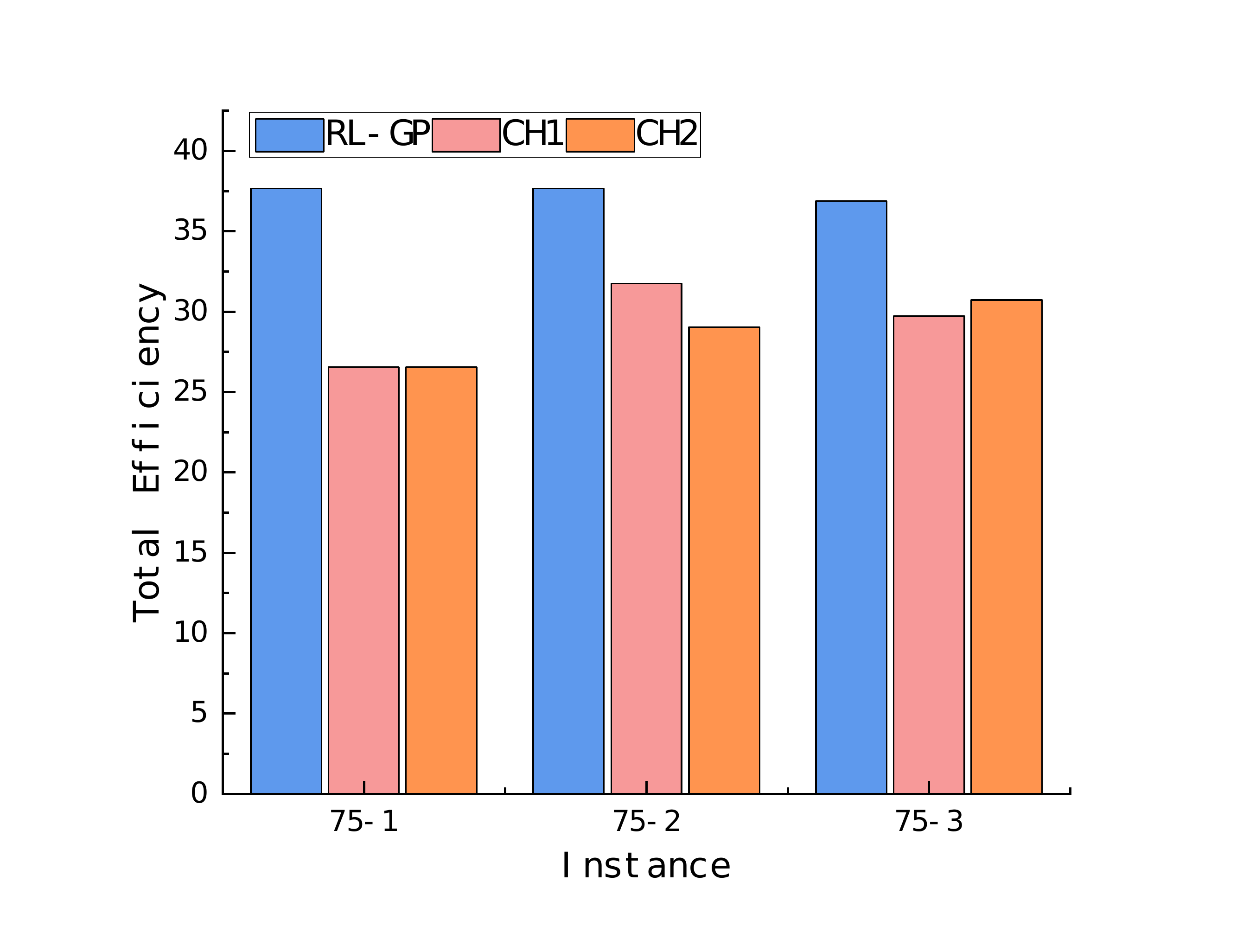}} \qquad
	\subfloat[Results for instances with 100 positions]{\includegraphics[width=.3\textwidth]{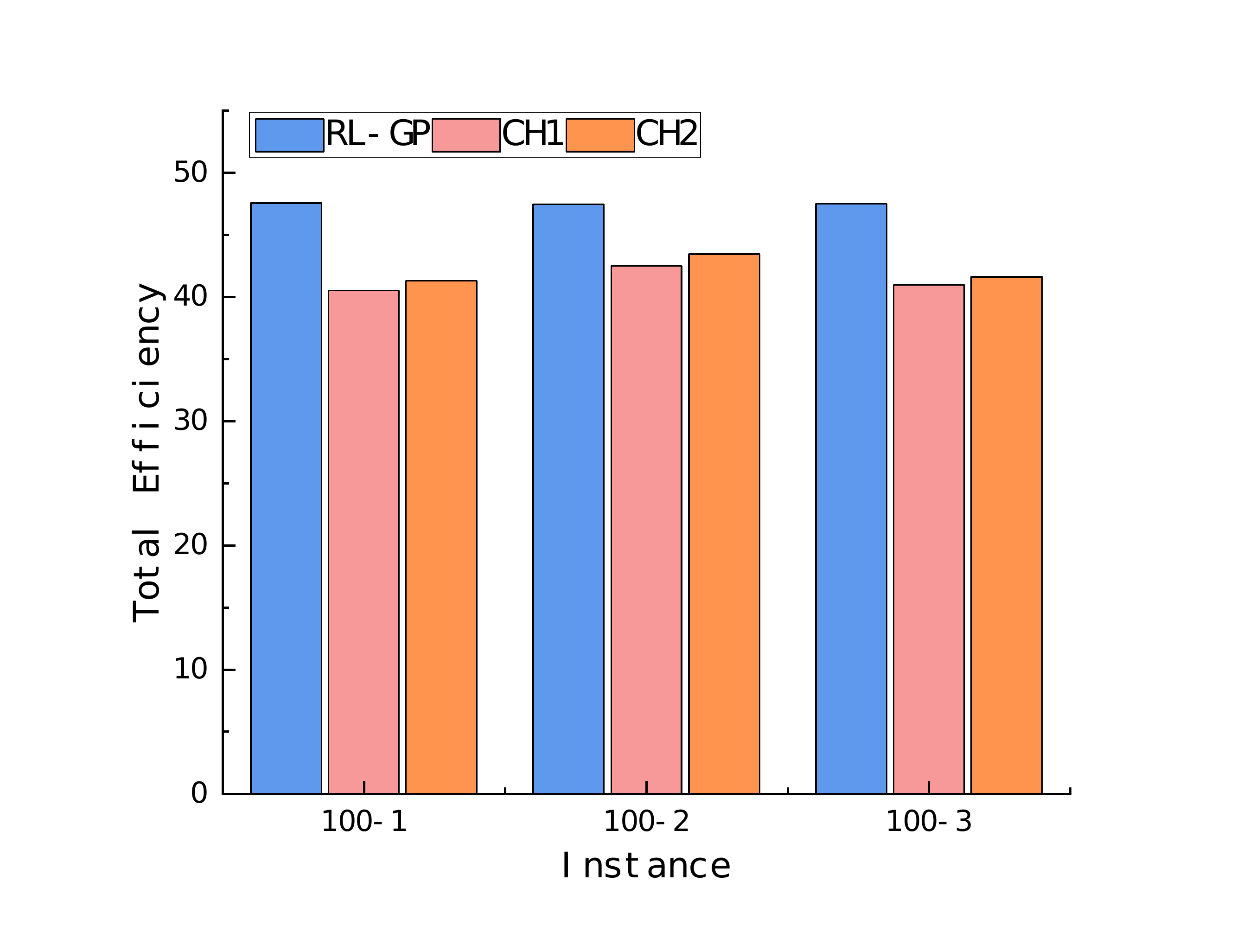}} \qquad
	\subfloat[Results for instances with 125 positions]{\includegraphics[width=.3\textwidth]{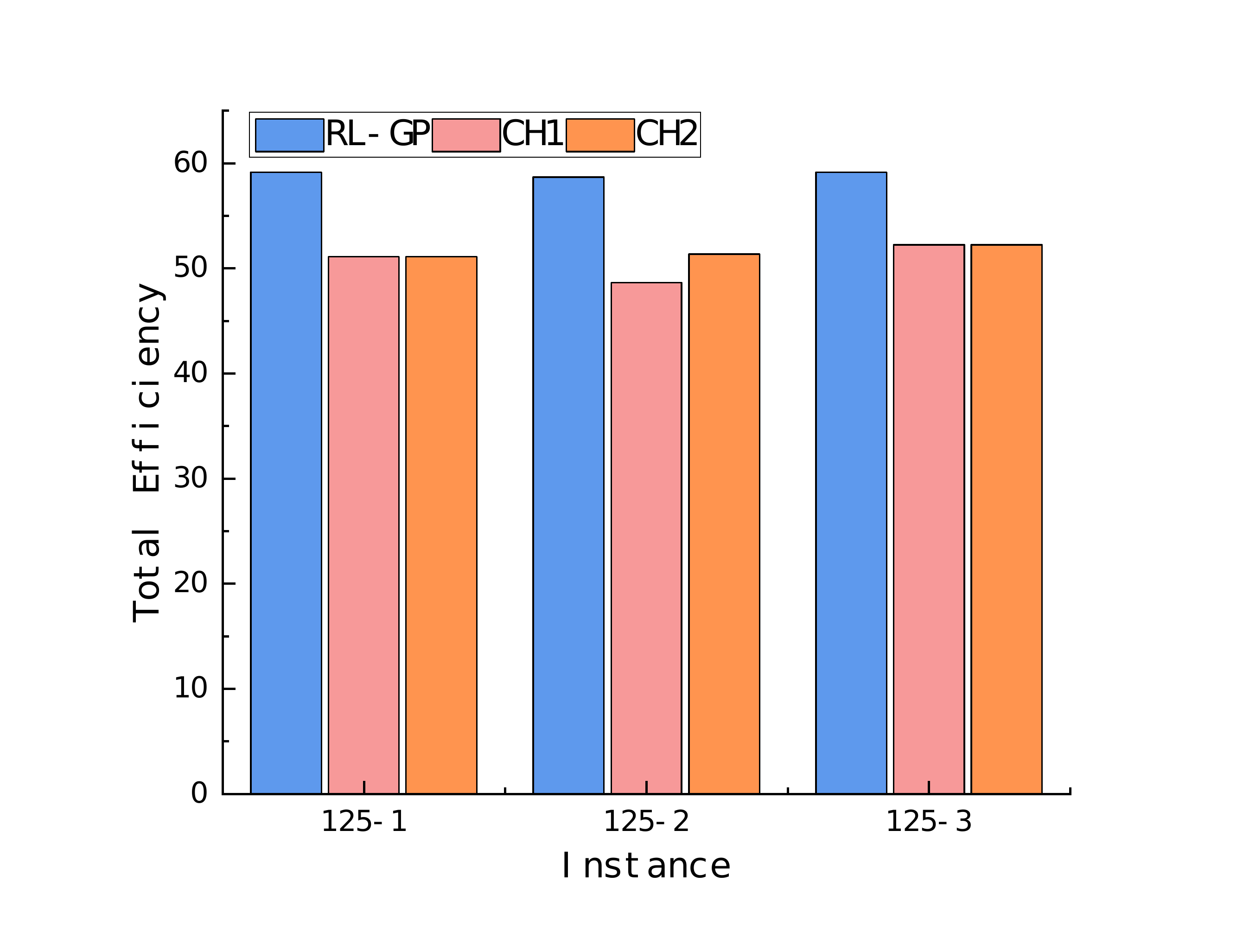}} \qquad
	\subfloat[Results for instances with 150 positions]{\includegraphics[width=.3\textwidth]{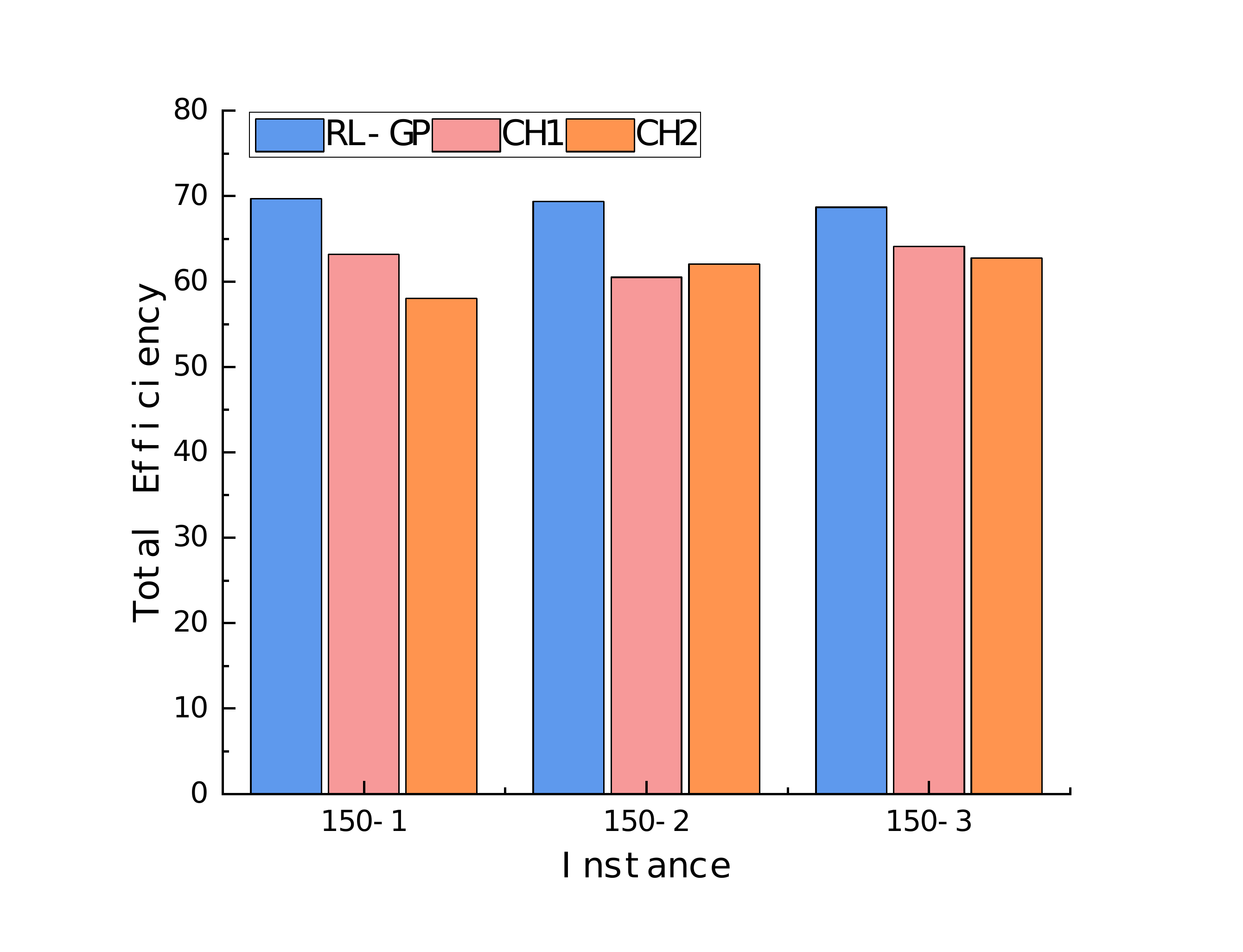}} \qquad
	\subfloat[Results for instances with 175 positions]{\includegraphics[width=.3\textwidth]{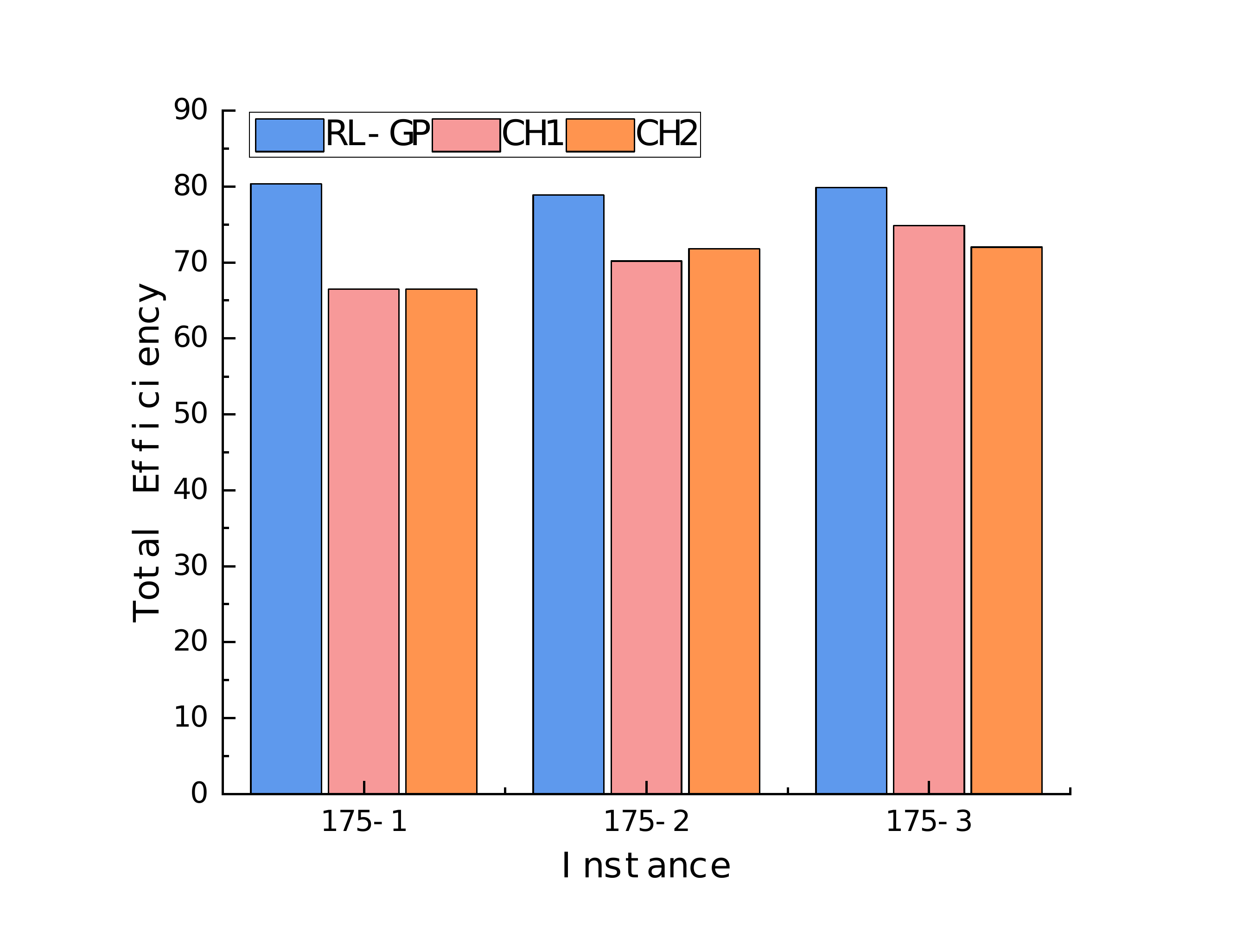}} \qquad
	\subfloat[Results for instances with 200 positions]{\includegraphics[width=.3\textwidth]{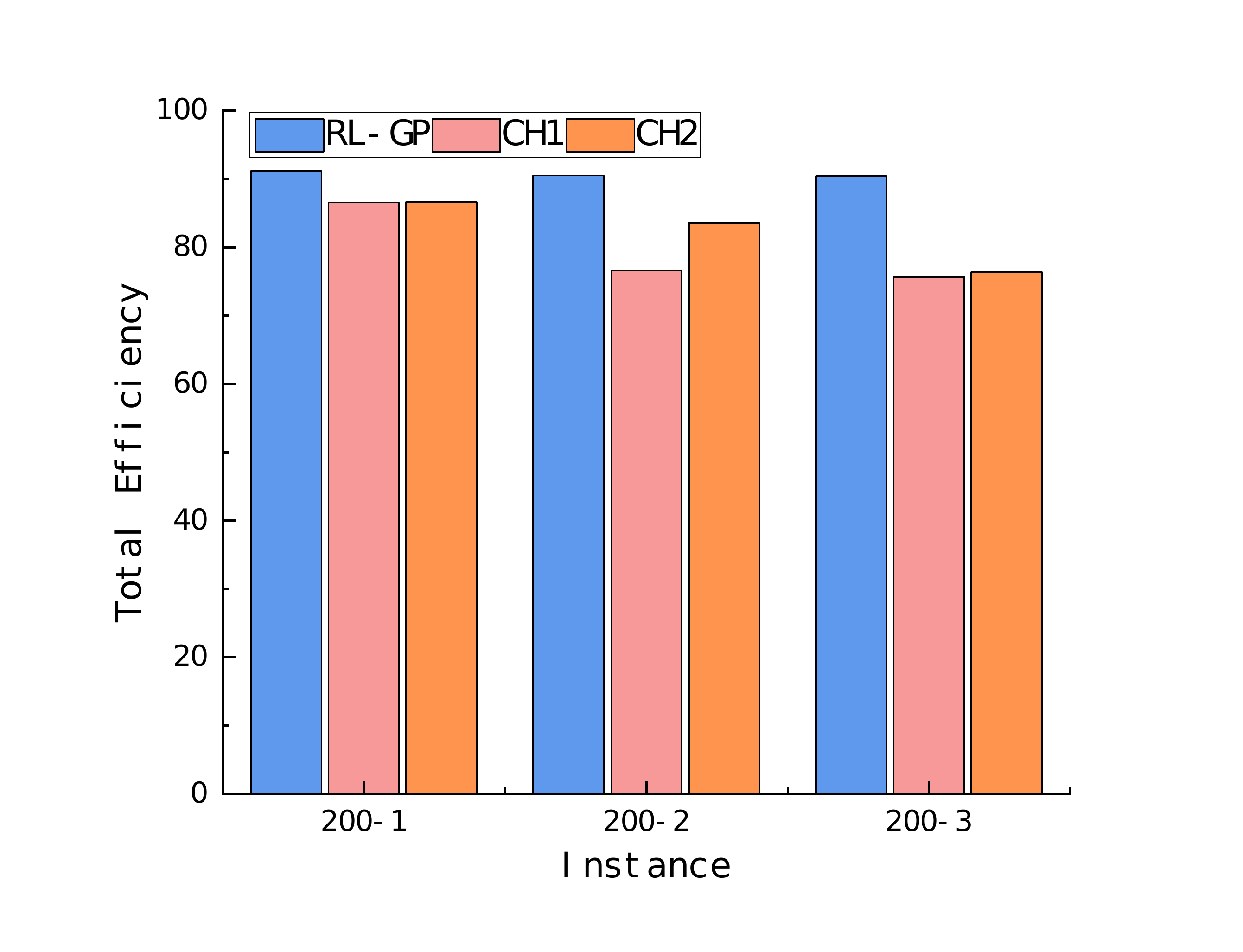}} \qquad
	\subfloat[Results for instances with 225 positions]{\includegraphics[width=.3\textwidth]{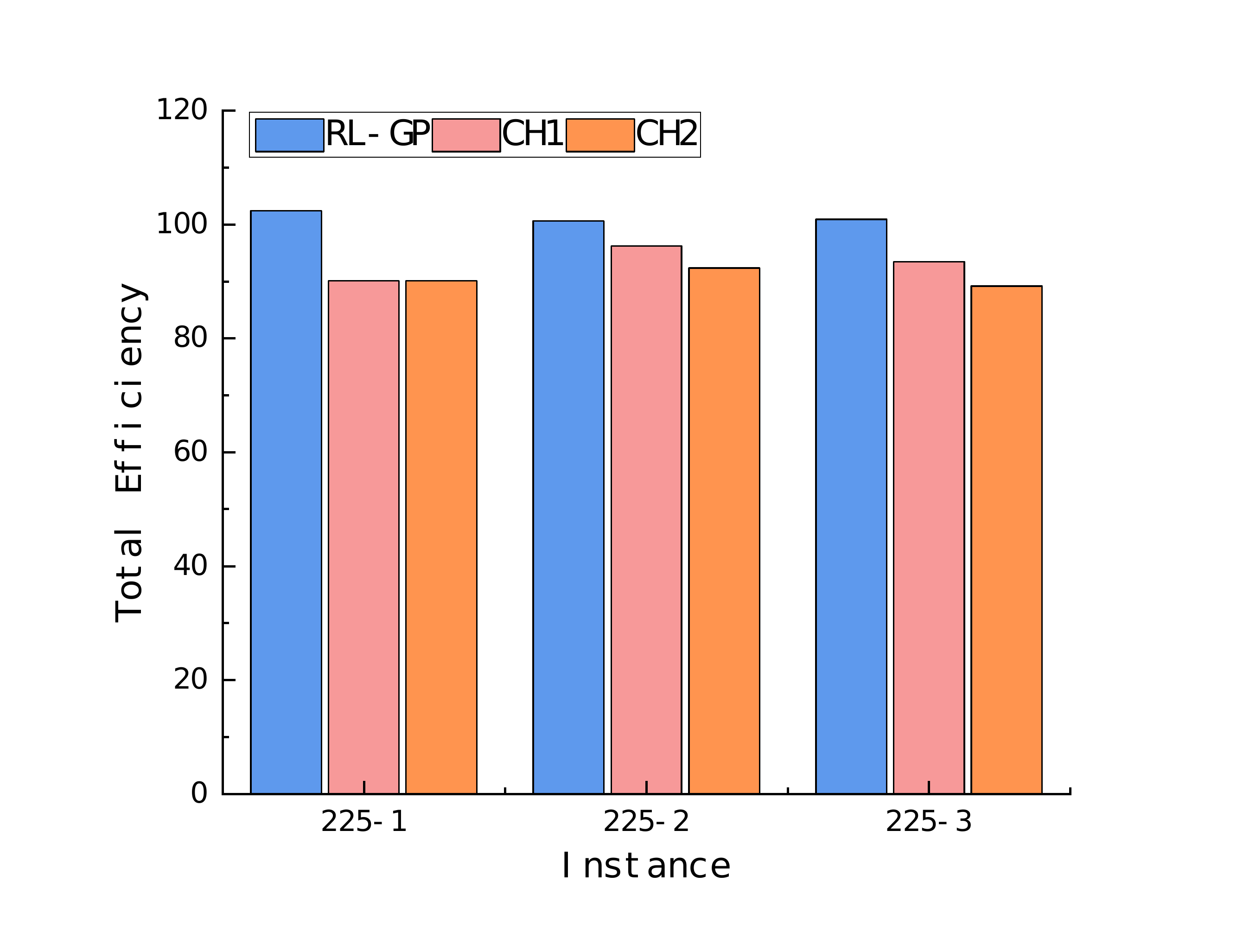}} 
	\caption{Comparison results of RL-GP with construction heuristic algorithms}
	\label{Comparison results of RL-GP with construction heuristic algorithms}
\end{figure*}

Evaluation metrics: For the learning process, the length of the learning time of the algorithms at different instance sizes can reflect the learning speed of the algorithms. The value of the objective function is used to evaluate the performance of the GP algorithm in the testing process. As for other search algorithms, the maximum value (denoted as Max) and the average value (denoted as Ave) of 10 runs are taken due to the uncertainty of the search process. In addition, the percentage of the difference between the proposed algorithm and the comparison algorithms (denoted as Gap) is also used to evaluate the performance of the algorithm in producing the hyper-heuristic rule.

\subsection{Experimental Results}

The experimental results contain comparisons between RL-GP and other GPs, construction heuristics algorithms, and other search algorithms, respectively. Then, the effect of the learned programming heuristic, the ratio between posts and skills on the performance of the algorithm is also analyzed.

\subsubsection{Comparisons with other GPs}
As an improved form of GP, whether the adopted improvement strategy can play a positive role in RL-GP performance in terms of both time and solution quality needs to be verified experimentally. First, the learning times used by different GP algorithms are compared. As shown in Figure \ref{Boxplots of the learning time for instances with different scale positions}, the learning time consumed by RL-GP is significantly shorter than that of the GP and GP-E algorithms, which reflects that the surrogate model used in the algorithm can significantly reduce the computational cost of evaluating individual fitness values. The usefulness of such an adaptation evaluation method is also more obvious as the problem size increases. For example, the number of candidates in a 100-job instance is twice as many as the number of candidates in a 50-job instance. Select a subset of these candidates to form a team, and the increased number of judgments of constraints will make each solution generation take longer. However, the use of the surrogate model to compute the fitness function values does not result in significant additional computation as the problem size increases. It is worth mentioning that the surrogate model calculation is not a complete substitute for the real solution evaluation method. Therefore, after a certain number of generations, RL-GP will calculate the real fitness function values and updates the optimal solution.

After that, the performance of RL-GP and two other comparable GP algorithms for problem-solving is compared. As shown in Table \ref{Comparison results of RL-GP with BGP and GP-E}, RL-GP can obtain a higher matching solution than the BGP and GP-E, where TE denotes the total efficiency. In RL-GP, the reinforcement learning method to select the population search mode can help the algorithm to adjust the search strategy according to the information obtained during the search. If the problem solution space is extremely complex, the combined search methods make it easier for the algorithm to balance exploration and exploitation. The use of elite individuals in the population is more likely to drive the search in a good direction. Additionally, the GP-E outperforms the BGP in the majority of cases.

\begin{figure}[htp]
	\centering
	\includegraphics[width=0.25\textwidth]{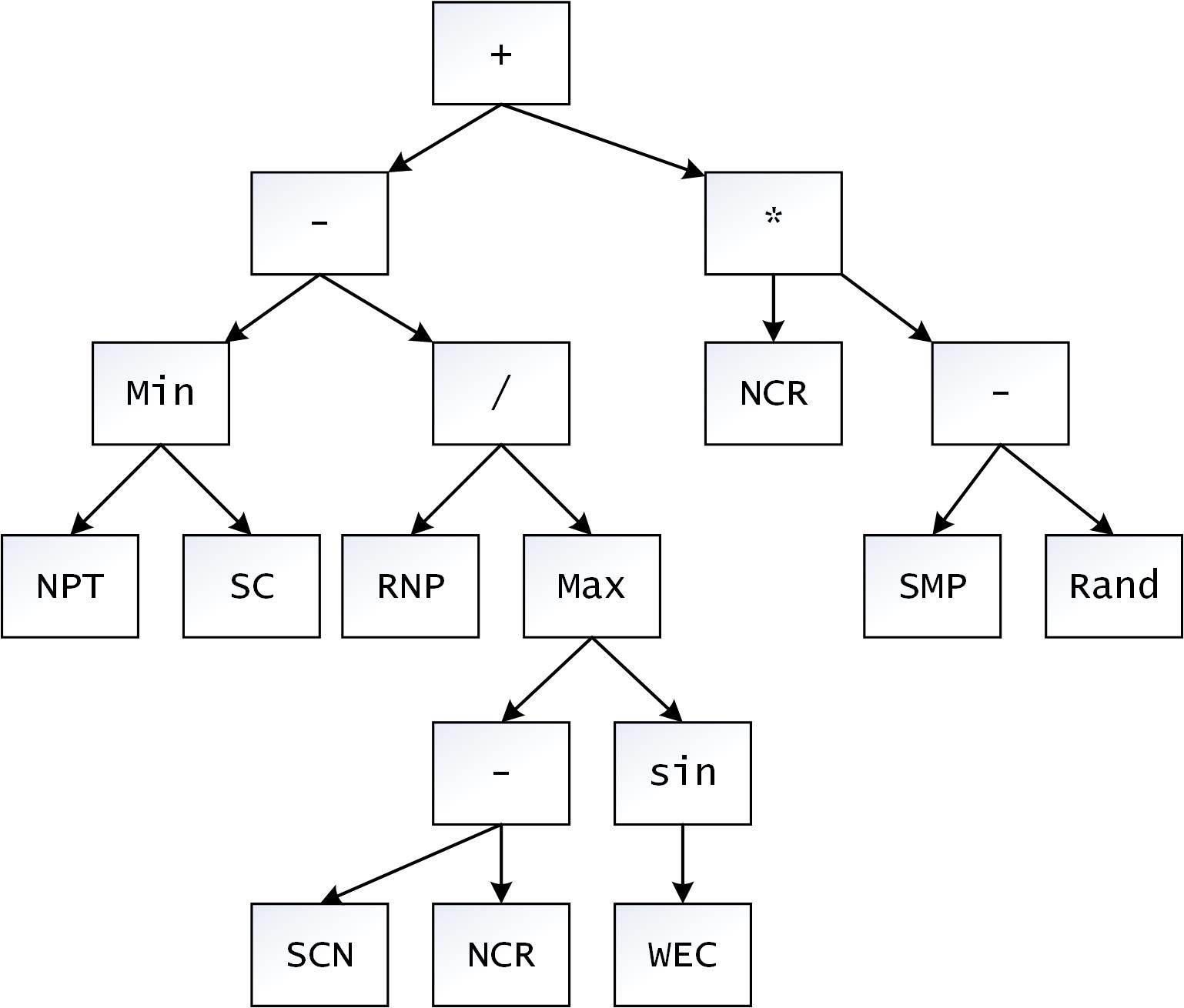}
	\caption{Good heuristic example of creating a team with 100 positions}
	\label{Good heuristic example of creating a team with 100 positions}
\end{figure}

\begin{figure}[htp]
	\centering
	\includegraphics[width=0.3\textwidth]{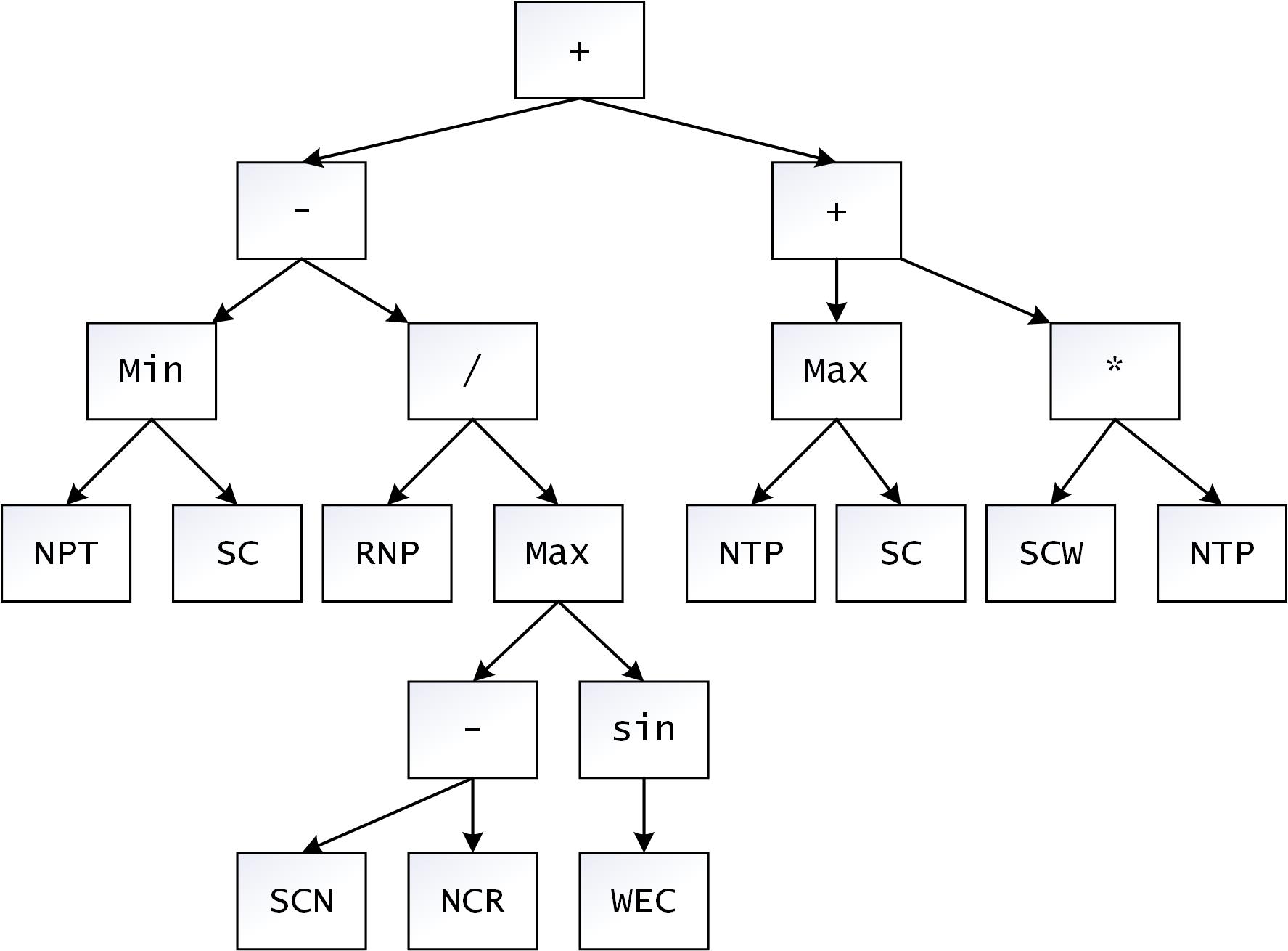}
	\caption{Good heuristic example of creating a team with 200 positions}
	\label{Good heuristic example of creating a team with 200 positions}
\end{figure}

\subsubsection{Comparisons with the construction heuristic algorithms}
To evaluate the performance of the heuristic rules derived from the training of the RL-GP algorithm in solving the TFP-PJM problem, two construction heuristics are used as benchmarks for comparison. Both the construction heuristic algorithms and the RL-GP algorithm can quickly obtain a team formation solution. This section focuses on evaluating the solution performance of the algorithms based on the overall efficiency of the solution. As shown in Figure \ref{Comparison results of RL-GP with construction heuristic algorithms}, the RL-GP algorithm can obtain team solutions that surpass those of the CH1 and CH2 algorithms. The obvious gap between the algorithms reflects the efficacy of population iterative search in finding the appropriate strategy. The use of the surrogate model to evaluate the fitness function values of RL-GP inevitably results in the omission of some individuals who can potentially obtain a high-matching solution. However, for complex combinatorial optimization problems like TFP-PJM, it is imperative to strike a balance between learning time and solution accuracy. Therefore, for solving such problems, there is no guarantee of an optimal solution. Instead, a high-quality solution that is not the best is equally acceptable.

\subsubsection{Comparisons with other search algorithms}
To further enhance the persuasiveness of the algorithm's solution performance, we also conduct a comparison with two search algorithms. As shown in Table \ref{Comparison results of RL-GP with IGA and VNS}, IGA can find efficient teams in some instances after spending a lot of time. Meanwhile, RL-GP outperforms the two search algorithms in certain instances, with equivalent or superior team formation solutions. Additionally, VNS performs worse than IGA. Through learning, RL-GP gains the heuristic of not only having the generalization ability but also building an efficient team. The learning process is made more efficient by the reinforcement learning method, where the state of each search mode determines the action to be used in the subsequent learning process. The information obtained by the algorithm during the learning process can guide the population to find the formation strategy that is more likely to obtain high-efficiency values.

\begin{table*}[ht]
	\small
	\caption{Comparison results of RL-GP with IGA and VNS}
	\label{Comparison results of RL-GP with IGA and VNS}
	\centering
	\begin{tabular}{llllllll}
		\toprule[1.5pt]
		\multicolumn{1}{c}{\multirow{2}{*}{Instance}} & \multicolumn{1}{c}{\multirow{2}{*}{RL-GP}} & IGA              &         &        & VNS     &         &          \\
		\cmidrule(r){3-5} \cmidrule(r){6-8}
		\multicolumn{1}{c}{} & \multicolumn{1}{c}{} & Best             & Ave     & Gap (\%)    & Best    & Ave     & Gap (\%)     \\
		\midrule[1pt]
		25-1  & \textbf{14.50}  & \textbf{14.50}  & 14.39 & 0.00  & 13.01 & 10.75 & -10.30 \\
		25-2  & \textbf{14.75}  & \textbf{14.75}  & 14.44 & 0.00  & 11.06 & 10.72 & -25.01 \\
		25-3  & \textbf{14.50}  & 13.86           & 13.70 & -4.43 & 12.30 & 11.56 & -15.14 \\
		50-1  & \textbf{25.02}  & 24.83           & 24.09 & -0.75 & 24.08 & 21.73 & -3.73  \\
		50-2  & 24.65           & \textbf{25.16}  & 24.21 & 2.04  & 23.48 & 22.61 & -4.78  \\
		50-3  & \textbf{25.02}  & 24.94           & 24.05 & -0.33 & 22.57 & 22.23 & -9.77  \\
		75-1  & 37.66           & \textbf{37.72}  & 36.54 & 0.15  & 32.82 & 30.87 & -12.86 \\
		75-2  & \textbf{37.66}  & 36.74           & 36.22 & -2.44 & 33.41 & 33.25 & -11.29 \\
		75-3  & \textbf{36.86}  & 36.74           & 35.70 & -0.35 & 33.32 & 32.59 & -9.61  \\
		100-1 & \textbf{47.55}  & 46.93           & 45.28 & -1.30 & 42.99 & 41.91 & -9.59  \\
		100-2 & \textbf{47.48}  & 46.89           & 45.21 & -1.24 & 44.99 & 41.71 & -5.25  \\
		100-3 & \textbf{47.51}  & 47.48           & 45.35 & -0.07 & 43.58 & 41.78 & -8.27  \\
		125-1 & \textbf{59.16}  & 59.08           & 57.67 & -0.15 & 54.52 & 53.94 & -7.84  \\
		125-2 & \textbf{58.68}  & \textbf{59.21}  & 57.17 & 0.90  & 51.80 & 50.07 & -11.74 \\
		125-3 & \textbf{59.16}  & 58.20           & 56.83 & -1.63 & 53.23 & 50.21 & -10.02 \\
		150-1 & \textbf{69.69}  & 68.29           & 67.43 & -2.02 & 63.89 & 61.70 & -8.33  \\
		150-2 & \textbf{69.38}  & 69.03           & 67.76 & -0.49 & 64.08 & 62.66 & -7.64  \\
		150-3 & \textbf{68.68}  & \textbf{69.73}  & 67.70 & 1.53  & 66.75 & 62.75 & -2.82  \\
		175-1 & \textbf{80.38}  & \textbf{80.38}  & 79.09 & 0.00  & 76.70 & 75.33 & -4.57  \\
		175-2 & \textbf{78.88}  & \textbf{78.94}  & 78.33 & 0.07  & 74.80 & 72.71 & -5.17  \\
		175-3 & \textbf{79.86}  & 79.72           & 78.53 & -0.18 & 74.58 & 72.02 & -6.61  \\
		200-1 & \textbf{91.18}  & 90.42           & 88.62 & -0.83 & 82.81 & 80.00 & -9.17  \\
		200-2 & \textbf{90.52}  & 89.78           & 88.89 & -0.81 & 84.08 & 83.50 & -7.11  \\
		200-3 & \textbf{90.42}  & \textbf{90.66}  & 88.95 & 0.26  & 84.56 & 82.16 & -6.48  \\
		225-1 & \textbf{102.42} & \textbf{102.42} & 99.33 & 0.00  & 90.58 & 89.31 & -11.56 \\
		225-2 & \textbf{100.66} & \textbf{100.99} & 99.74 & 0.33  & 95.70 & 93.10 & -4.93  \\
		225-3 & \textbf{100.94} & 100.93          & 99.92 & -0.02 & 94.24 & 92.55 & -6.64 \\
		\bottomrule[1.5pt]
	\end{tabular}
\end{table*}

\begin{figure*}[htbp]
	\centering
	\subfloat[Results for instances with 25 positions]{\includegraphics[width=.3\textwidth]{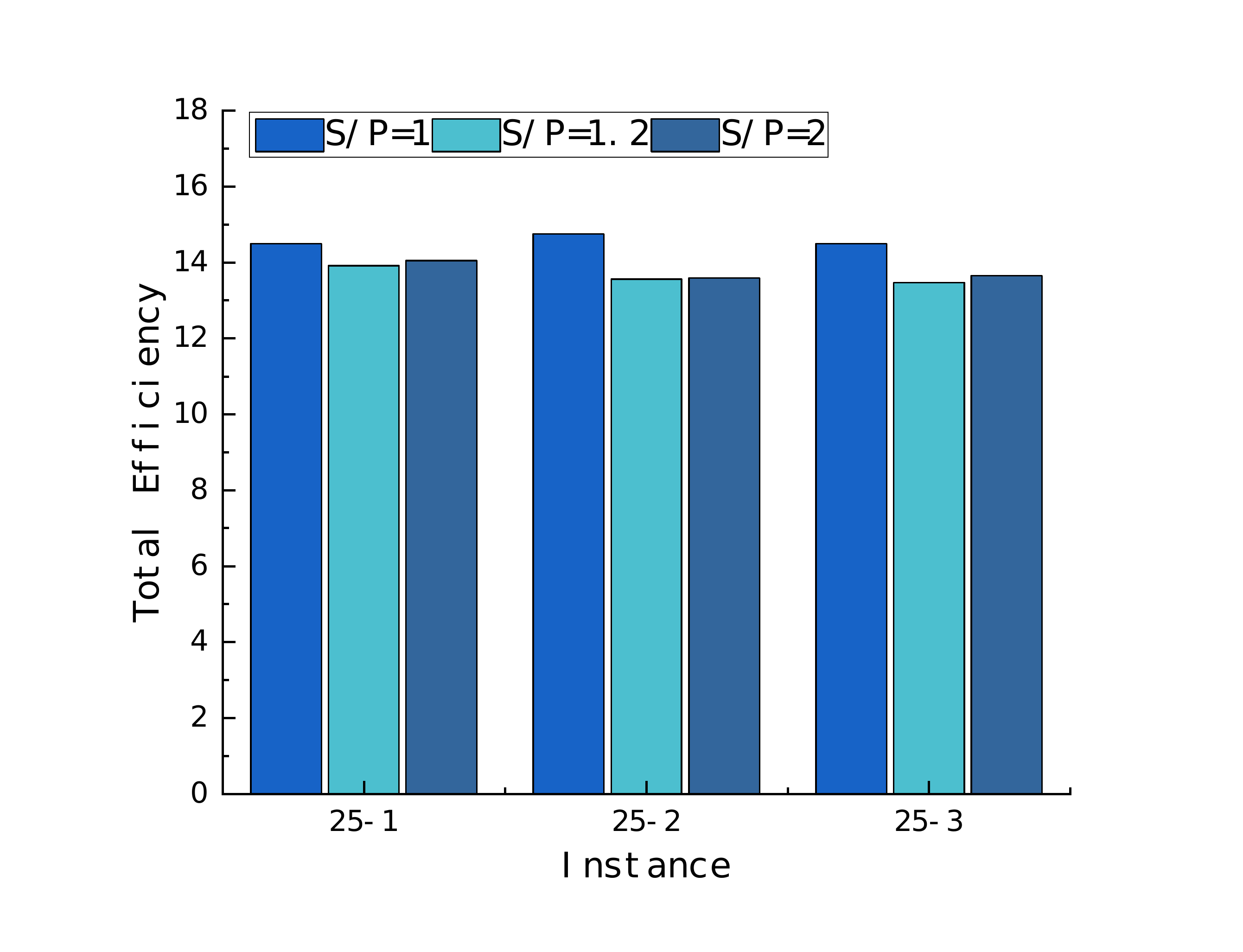}} \qquad
	\subfloat[Results for instances with 50 positions]{\includegraphics[width=.3\textwidth]{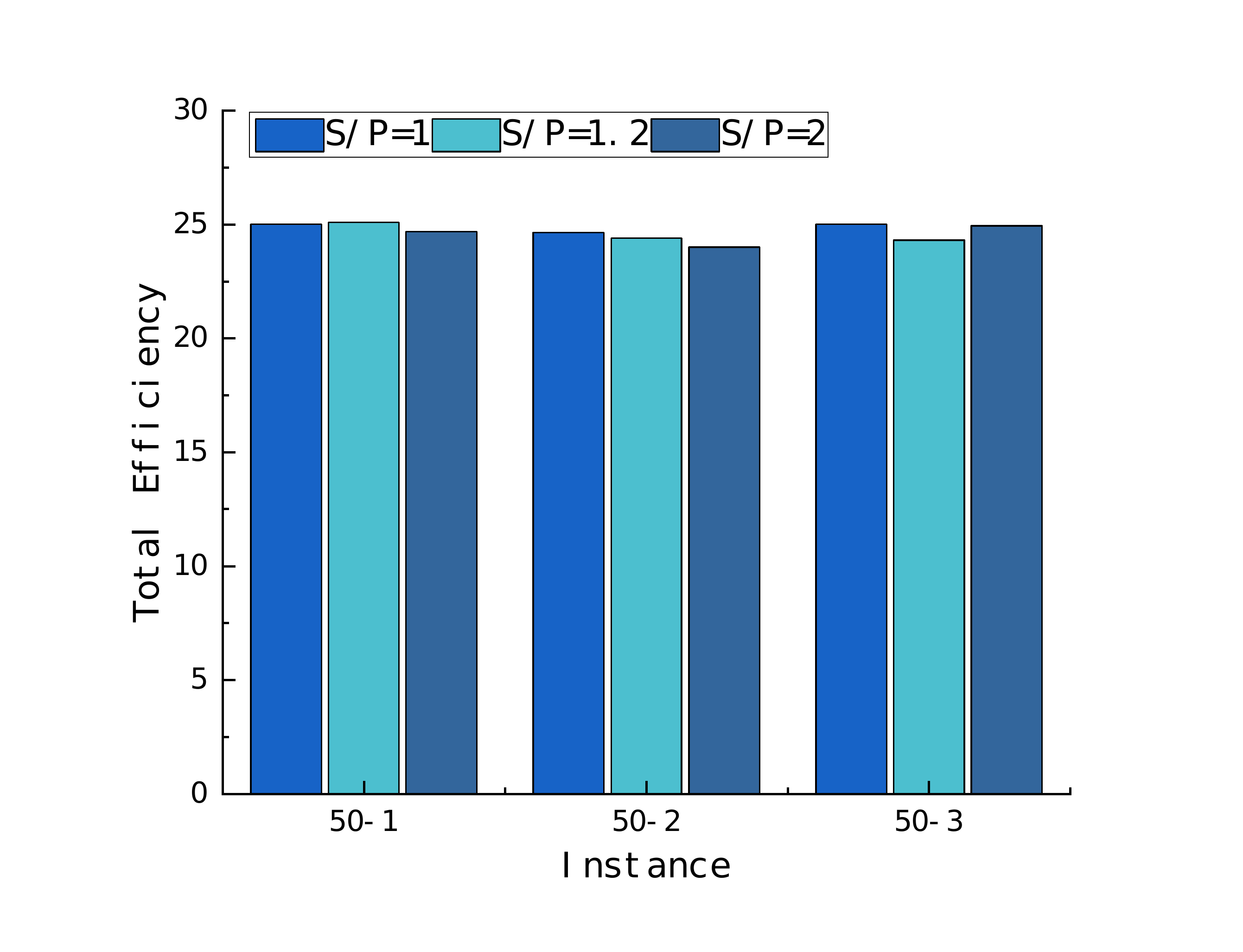}} \qquad
	\subfloat[Results for instances with 75 positions]{\includegraphics[width=.3\textwidth]{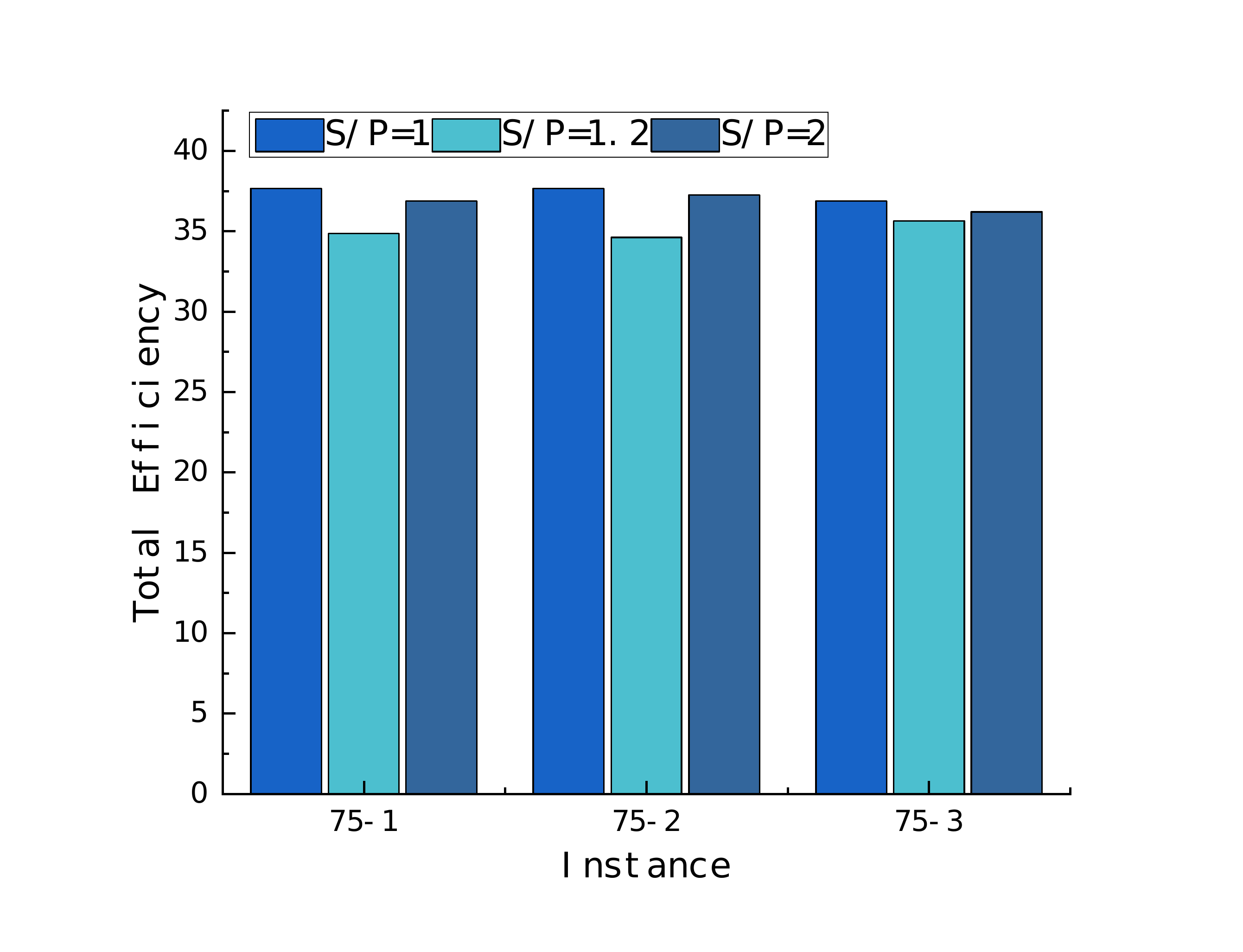}} 
	\caption{Results of the influence of different S/P ratios}
	\label{Results of the influence of different S/P ratios}
\end{figure*}

\subsubsection{Learned scheduling heuristics}
This section presents some learned heuristic rules for instances of size 100 and 200. Figure \ref{Good heuristic example of creating a team with 100 positions} and Figure \ref{Good heuristic example of creating a team with 200 positions} represent the well-performing heuristic for forming teams of 100 and 200 posts, respectively. It can be seen that the heuristics that generate efficient team solutions at two instance sizes exist partly similar structures. This shows that the agent tends to adopt similar actions during the search process based on the state information, and the individual evolution conducted in this manner also moves in the direction of being close to each other owing to the selected actions. The optimal heuristic for a large problem size is more complex, as the individual structures with more terminals and functions easily represent useful information about the candidates. The priority ranking generated in this way enables the construction of efficient teams by selecting suitable candidates from a vast pool of options.

\subsubsection{Effect of the ratio between skills and positions on algorithm performance}
We also analyze the impact of the proportional relationship between the number of different skill requirements and the number of positions on the efficiency values of the algorithm-generated solutions. Figure \ref{Results of the influence of different S/P ratios} shows the comparison of team efficiency results for multiple instances with different number ratios skills to positions (denoted as S/P). The ratio of skills to positions affects the difficulty of solving the problem to some extent. A higher ratio of skills to positions means more selectivity in the choice of candidates available for team formation, leading to a larger solution space for the problem. The difficulty of selecting candidates who meet the requirements increases accordingly.

\subsubsection{Discussion}
The above experiments comprehensively evaluate the performance of the proposed algorithm in solving TFP-PJM problems, using various GP algorithms, construction heuristic algorithms, and search algorithms for comparison. For example, the learning time of the RL-GP algorithm is less than half that of other GP algorithms, and the efficiency of the solution obtained by the RL-GP algorithm is also improved when compared with the two construction heuristics. In conclusion, the heuristic can be used to aid decision-making in generating team formation solutions in practical problem scenarios.  However, the randomness of the search can make a big difference in the depth of the generated high-quality individuals. In general, the heuristic with small depth has a strong generalization ability, but may not necessarily yield the most efficient teams. Conversely, the heuristic with large depth tends to obtain good planning performance in individual instances, but may not be effective in dealing with all situations. Therefore, the effect of the depth and frequency of the same terminal or function in the structure of GP-generated individuals on the results should be further analyzed in further studies. Additionally, some effective information about individuals can be directly applied to other individuals' learning strategies through transfer, which can also be a promising direction for future research.

\section{Conclusion}
When using reinforcement learning-assisted genetic programming to solve the team formation problem, we propose a population ensemble strategy that includes four search modes, and use reinforcement learning methods for search pattern selection, balancing the exploration and exploitation of population search. In addition, a K-Nearest Neighbor surrogate model is used in order to quickly evaluate the performance of iterative search for newly generated populations. The surrogate model is updated online based on the acquired information during the rule learning process to improve the model prediction accuracy. Extensive experiments have demonstrated that RL-GP requires significantly less time, between $1/5$ to $1/10$ of the time taken by other GP methods, to complete the learning process. Moreover, the obtained team formation rules generated by RL-GP outperform the constructed heuristic algorithm used for comparison in the experiments. In addition, some experiments show that RL-GP can generate team formation solutions that are either equivalent to or superior to those generated by the genetic algorithm and variable neighborhood search algorithm.

In the future, we will explore more complex team formation problems, such as multi-team formation and online adjustment of team formation schemes, and the constraints in the model will consider more factors. Furthermore, it is also worthwhile to explore the genetic programming algorithm combined with other methods such as data mining, ensemble learning, and transfer learning in further depth. This will further improve the performance and applicability of the genetic programming algorithm and effectively deal with more complex combinatorial optimization problems.



\bibliographystyle{IEEEtran}{}
\bibliography{mybib}

\begin{thebibliography}{10}
\providecommand{\url}[1]{#1}
\csname url@samestyle\endcsname
\providecommand{\newblock}{\relax}
\providecommand{\bibinfo}[2]{#2}
\providecommand{\BIBentrySTDinterwordspacing}{\spaceskip=0pt\relax}
\providecommand{\BIBentryALTinterwordstretchfactor}{4}
\providecommand{\BIBentryALTinterwordspacing}{\spaceskip=\fontdimen2\font plus
\BIBentryALTinterwordstretchfactor\fontdimen3\font minus
  \fontdimen4\font\relax}
\providecommand{\BIBforeignlanguage}[2]{{%
\expandafter\ifx\csname l@#1\endcsname\relax
\typeout{** WARNING: IEEEtran.bst: No hyphenation pattern has been}%
\typeout{** loaded for the language `#1'. Using the pattern for}%
\typeout{** the default language instead.}%
\else
\language=\csname l@#1\endcsname
\fi
#2}}
\providecommand{\BIBdecl}{\relax}
\BIBdecl

\bibitem{liu2021data}
J.~Liu, J.~Huang, T.~Wang, L.~Xing, and R.~He, ``A data-driven analysis of
  employee development based on working expertise,'' \emph{IEEE Transactions on
  Computational Social Systems}, vol.~8, no.~2, pp. 410--422, 2021.

\bibitem{he2021self}
M.~He, D.~Shen, T.~Wang, H.~Zhao, Z.~Zhang, and R.~He, ``Self-attentional
  multi-field features representation and interaction learning for person-job
  fit,'' \emph{IEEE Transactions on Computational Social Systems}, 2021.

\bibitem{juarez2021comprehensive}
J.~Ju{\'a}rez, C.~Santos, and C.~A. Brizuela, ``A comprehensive review and a
  taxonomy proposal of team formation problems,'' \emph{ACM Computing Surveys
  (CSUR)}, vol.~54, no.~7, pp. 1--33, 2021.

\bibitem{ahmed2013multi}
F.~Ahmed, K.~Deb, and A.~Jindal, ``Multi-objective optimization and decision
  making approaches to cricket team selection,'' \emph{Applied Soft Computing},
  vol.~13, no.~1, pp. 402--414, 2013.

\bibitem{abdelsalam2009multi}
H.~M. Abdelsalam, ``Multi-objective team forming optimization for integrated
  product development projects,'' \emph{Foundations of Computational
  Intelligence Volume 3: Global Optimization}, pp. 461--478, 2009.

\bibitem{niveditha2017genetic}
M.~Niveditha, G.~Swetha, U.~Poornima, and R.~Senthilkumar, ``A genetic approach
  for tri-objective optimization in team formation,'' in \emph{2016 Eighth
  International Conference on Advanced Computing (ICoAC)}.\hskip 1em plus 0.5em
  minus 0.4em\relax IEEE, 2017, pp. 123--130.

\bibitem{perez2019players}
M.~{\'A}. P{\'e}rez-Toledano, F.~J. Rodriguez, J.~Garc{\'\i}a-Rubio, and S.~J.
  Iba{\~n}ez, ``Players’ selection for basketball teams, through performance
  index rating, using multiobjective evolutionary algorithms,'' \emph{PloS
  one}, vol.~14, no.~9, p. e0221258, 2019.

\bibitem{okimoto2016mission}
T.~Okimoto, T.~Ribeiro, D.~Bouchabou, and K.~Inoue, ``Mission oriented robust
  multi-team formation and its application to robot rescue simulation,'' in
  \emph{Twenty-Fifth International Joint Conference on Artificial Intelligence
  (IJCAI-16)}, 2016.

\bibitem{liemhetcharat2012modeling}
S.~Liemhetcharat and M.~Veloso, ``Modeling and learning synergy for team
  formation with heterogeneous agents,'' in \emph{Proceedings of the 11th
  International Conference on Autonomous Agents and Multiagent Systems-Volume
  1}, 2012, pp. 365--374.

\bibitem{wi2009team}
H.~Wi, S.~Oh, J.~Mun, and M.~Jung, ``A team formation model based on knowledge
  and collaboration,'' \emph{Expert Systems with Applications}, vol.~36, no.~5,
  pp. 9121--9134, 2009.

\bibitem{zhang2013multi}
L.~Zhang and X.~Zhang, ``Multi-objective team formation optimization for new
  product development,'' \emph{Computers \& Industrial Engineering}, vol.~64,
  no.~3, pp. 804--811, 2013.

\bibitem{gutierrez2016multiple}
J.~H. Guti{\'e}rrez, C.~A. Astudillo, P.~Ballesteros-P{\'e}rez,
  D.~Mora-Meli{\`a}, and A.~Candia-V{\'e}jar, ``The multiple team formation
  problem using sociometry,'' \emph{Computers \& Operations Research}, vol.~75,
  pp. 150--162, 2016.

\bibitem{zhang2021multitask}
F.~Zhang, Y.~Mei, S.~Nguyen, K.~C. Tan, and M.~Zhang, ``Multitask genetic
  programming-based generative hyperheuristics: A case study in dynamic
  scheduling,'' \emph{IEEE Transactions on Cybernetics}, vol.~52, no.~10, pp.
  10\,515--10\,528, 2021.

\bibitem{ardeh2021genetic}
M.~A. Ardeh, Y.~Mei, and M.~Zhang, ``Genetic programming with knowledge
  transfer and guided search for uncertain capacitated arc routing problem,''
  \emph{IEEE Transactions on Evolutionary Computation}, vol.~26, no.~4, pp.
  765--779, 2021.

\bibitem{tan2020cooperative}
B.~Tan, H.~Ma, Y.~Mei, and M.~Zhang, ``A cooperative coevolution genetic
  programming hyper-heuristics approach for on-line resource allocation in
  container-based clouds,'' \emph{IEEE Transactions on Cloud Computing},
  vol.~10, no.~3, pp. 1500--1514, 2020.

\bibitem{boon2003team}
B.~H. Boon and G.~Sierksma, ``Team formation: Matching quality supply and
  quality demand,'' \emph{European Journal of Operational Research}, vol. 148,
  no.~2, pp. 277--292, 2003.

\bibitem{fitzpatrick2005forming}
E.~L. Fitzpatrick and R.~G. Askin, ``Forming effective worker teams with
  multi-functional skill requirements,'' \emph{Computers \& Industrial
  Engineering}, vol.~48, no.~3, pp. 593--608, 2005.

\bibitem{xu2012modeling}
Y.~Xu, J.~Ma, and X.~Guo, ``Modeling researchers’ characteristics for the
  formation of research team,'' 2012.

\bibitem{baykasoglu2007project}
A.~Baykasoglu, T.~Dereli, and S.~Das, ``Project team selection using fuzzy
  optimization approach,'' \emph{Cybernetics and Systems: An International
  Journal}, vol.~38, no.~2, pp. 155--185, 2007.

\bibitem{wang2015comparative}
X.~Wang, Z.~Zhao, and W.~Ng, ``A comparative study of team formation in social
  networks,'' in \emph{Database Systems for Advanced Applications: 20th
  International Conference, DASFAA 2015, Hanoi, Vietnam, April 20-23, 2015,
  Proceedings, Part I 20}.\hskip 1em plus 0.5em minus 0.4em\relax Springer,
  2015, pp. 389--404.

\bibitem{strnad2010fuzzy}
D.~Strnad and N.~Guid, ``A fuzzy-genetic decision support system for project
  team formation,'' \emph{Applied Soft Computing}, vol.~10, no.~4, pp.
  1178--1187, 2010.

\bibitem{bhowmik2014submodularity}
A.~Bhowmik, V.~Borkar, D.~Garg, and M.~Pallan, ``Submodularity in team
  formation problem,'' in \emph{Proceedings of the 2014 SIAM international
  conference on data mining}.\hskip 1em plus 0.5em minus 0.4em\relax SIAM,
  2014, pp. 893--901.

\bibitem{xu2021genetic}
B.~Xu, Y.~Mei, Y.~Wang, Z.~Ji, and M.~Zhang, ``Genetic programming with delayed
  routing for multiobjective dynamic flexible job shop scheduling,''
  \emph{Evolutionary Computation}, vol.~29, no.~1, pp. 75--105, 2021.

\bibitem{wang2021genetic}
S.~Wang, Y.~Mei, M.~Zhang, and X.~Yao, ``Genetic programming with niching for
  uncertain capacitated arc routing problem,'' \emph{IEEE Transactions on
  Evolutionary Computation}, vol.~26, no.~1, pp. 73--87, 2021.

\bibitem{zhang2021surrogate}
F.~Zhang, Y.~Mei, S.~Nguyen, M.~Zhang, and K.~C. Tan, ``Surrogate-assisted
  evolutionary multitask genetic programming for dynamic flexible job shop
  scheduling,'' \emph{IEEE Transactions on Evolutionary Computation}, vol.~25,
  no.~4, pp. 651--665, 2021.

\bibitem{farasat2016social}
A.~Farasat and A.~G. Nikolaev, ``Social structure optimization in team
  formation,'' \emph{Computers \& Operations Research}, vol.~74, pp. 127--142,
  2016.

\bibitem{fathian2017new}
M.~Fathian, M.~Saei-Shahi, and A.~Makui, ``A new optimization model for
  reliable team formation problem considering experts’ collaboration
  network,'' \emph{IEEE Transactions on Engineering Management}, vol.~64,
  no.~4, pp. 586--593, 2017.

\bibitem{xu2007intuitionistic}
Z.~Xu, ``Intuitionistic fuzzy aggregation operators,'' \emph{IEEE Transactions
  on Fuzzy Systems}, vol.~15, no.~6, pp. 1179--1187, 2007.

\bibitem{he2016intuitionistic}
X.~He, Y.~Wu, and D.~Yu, ``Intuitionistic fuzzy multi-criteria decision making
  with application to job hunting: A comparative perspective,'' \emph{Journal
  of Intelligent \& Fuzzy Systems}, vol.~30, no.~4, pp. 1935--1946, 2016.

\bibitem{mei2022explainable}
Y.~Mei, Q.~Chen, A.~Lensen, B.~Xue, and M.~Zhang, ``Explainable artificial
  intelligence by genetic programming: A survey,'' \emph{IEEE Transactions on
  Evolutionary Computation}, 2022.

\bibitem{kallestad2023general}
J.~Kallestad, R.~Hasibi, A.~Hemmati, and K.~S{\"o}rensen, ``A general deep
  reinforcement learning hyperheuristic framework for solving combinatorial
  optimization problems,'' \emph{European Journal of Operational Research},
  2023.

\bibitem{hildebrandt2015using}
T.~Hildebrandt and J.~Branke, ``On using surrogates with genetic programming,''
  \emph{Evolutionary Computation}, vol.~23, no.~3, pp. 343--367, 2015.

\bibitem{espejo2009survey}
P.~G. Espejo, S.~Ventura, and F.~Herrera, ``A survey on the application of
  genetic programming to classification,'' \emph{IEEE Transactions on Systems,
  Man, and Cybernetics, Part C (Applications and Reviews)}, vol.~40, no.~2, pp.
  121--144, 2009.

\bibitem{dabbaghjamanesh2020reinforcement}
M.~Dabbaghjamanesh, A.~Moeini, and A.~Kavousi-Fard, ``Reinforcement
  learning-based load forecasting of electric vehicle charging station using
  q-learning technique,'' \emph{IEEE Transactions on Industrial Informatics},
  vol.~17, no.~6, pp. 4229--4237, 2020.

\bibitem{da2018parallel}
L.~M. Da~Silva, M.~F. Torquato, and M.~A. Fernandes, ``Parallel implementation
  of reinforcement learning q-learning technique for fpga,'' \emph{IEEE
  Access}, vol.~7, pp. 2782--2798, 2018.

\bibitem{doltsinis2014mdp}
S.~Doltsinis, P.~Ferreira, and N.~Lohse, ``An mdp model-based reinforcement
  learning approach for production station ramp-up optimization: Q-learning
  analysis,'' \emph{IEEE Transactions on Systems, Man, and Cybernetics:
  Systems}, vol.~44, no.~9, pp. 1125--1138, 2014.

\bibitem{moon2020generalized}
J.~Moon, ``Generalized risk-sensitive optimal control and
  hamilton--jacobi--bellman equation,'' \emph{IEEE Transactions on Automatic
  Control}, vol.~66, no.~5, pp. 2319--2325, 2020.

\bibitem{ye2017novel}
W.~Ye, W.~Feng, and S.~Fan, ``A novel multi-swarm particle swarm optimization
  with dynamic learning strategy,'' \emph{Applied Soft Computing}, vol.~61, pp.
  832--843, 2017.

\bibitem{wu2019ensemble}
G.~Wu, R.~Mallipeddi, and P.~N. Suganthan, ``Ensemble strategies for
  population-based optimization algorithms--a survey,'' \emph{Swarm and
  Evolutionary Computation}, vol.~44, pp. 695--711, 2019.

\bibitem{zhang2017efficient}
S.~Zhang, X.~Li, M.~Zong, X.~Zhu, and R.~Wang, ``Efficient knn classification
  with different numbers of nearest neighbors,'' \emph{IEEE Transactions on
  Neural Networks and Learning Systems}, vol.~29, no.~5, pp. 1774--1785, 2017.

\bibitem{wang2022improved}
H.~Wang, J.~Li, Y.~Song, J.~Huang, J.~Li, and Y.~Chen, ``An improved genetic
  algorithm for team formation problem,'' in \emph{2022 IEEE Symposium Series
  on Computational Intelligence (SSCI)}.\hskip 1em plus 0.5em minus 0.4em\relax
  IEEE, 2022, pp. 774--781.

\bibitem{hansen2019variable}
P.~Hansen, N.~Mladenovi{\'c}, J.~Brimberg, and J.~A.~M. P{\'e}rez,
  \emph{Variable neighborhood search}.\hskip 1em plus 0.5em minus 0.4em\relax
  Springer, 2019.

\end{thebibliography}

\end{document}